\documentclass[11pt,a4paper]{article}
\pdfoutput=1
\usepackage{amsmath}
\usepackage[T1]{fontenc}
\usepackage[utf8]{inputenc}
\usepackage{newtxtext}
\usepackage{newtxmath}
\usepackage{newunicodechar}
\newunicodechar{±}{\ensuremath{\pm}}
\newunicodechar{×}{\ensuremath{\times}}
\newunicodechar{Δ}{\ensuremath{\Delta}}
\newunicodechar{σ}{\ensuremath{\sigma}}
\newunicodechar{†}{\textdagger{}}
\newunicodechar{→}{\ensuremath{\rightarrow}}
\newunicodechar{−}{\ensuremath{-}}
\newunicodechar{≥}{\ensuremath{\geq}}

\usepackage{dsfont}
\usepackage[margin=2.4cm]{geometry}
\usepackage{graphicx,booktabs,tabularx,array,rotating}
\usepackage{placeins}
\usepackage{caption}
\DeclareCaptionLabelSeparator{pipe}{\ $|$\ }
\usepackage{microtype}
\usepackage{setspace}
\usepackage[super,sort&compress,comma]{natbib}
\usepackage{xcolor}
\definecolor{brand}{HTML}{0E9C7E}\definecolor{brandink}{HTML}{0B5E50}\definecolor{brandbg}{HTML}{F1FBF8}
\usepackage{xurl}
\usepackage[colorlinks=true,linkcolor=brandink,citecolor=brandink,urlcolor=brandink]{hyperref}
\usepackage{titlesec}
\titleformat{\section}{\sffamily\large\bfseries\color{brandink}}{\thesection}{0.6em}{}
\titleformat{\subsection}{\sffamily\normalsize\bfseries\color{brandink}}{\thesubsection}{0.6em}{}
\titleformat{\subsubsection}{\normalfont\normalsize\bfseries}{}{0em}{}
\titlespacing*{\subsubsection}{0pt}{1.0ex plus .3ex}{0.2ex}
\titlespacing*{\section}{0pt}{1.6ex plus .4ex}{0.8ex}
\titlespacing*{\subsection}{0pt}{1.2ex plus .3ex}{0.5ex}
\usepackage{fancyhdr}
\makeatletter\renewcommand{\@biblabel}[1]{#1.}\makeatother
\newsavebox{\absbox}
\renewenvironment{abstract}{\par\noindent\setlength{\fboxsep}{9pt}\setlength{\fboxrule}{0.8pt}\begin{lrbox}{\absbox}\begin{minipage}{\dimexpr\linewidth-2\fboxsep-2\fboxrule\relax}\vspace{0.4ex}\centerline{\sffamily\large\bfseries\color{brandink}\abstractname}\vspace{0.6ex}\small\setstretch{1.08}\noindent\ignorespaces}{\par\vspace{0.4ex}\end{minipage}\end{lrbox}\fcolorbox{brand}{brandbg}{\usebox{\absbox}}\par\vspace{1.2em}}

\hypersetup{pdftitle={Operon-S1: a fast decision controller for long-horizon biomedical agents},pdfauthor={Yuejie Hou, Yueyuxiao Yang, Weiting Tang, Meng Yang}}

\begin{document}

\thispagestyle{plain}

\noindent\includegraphics[height=15mm]{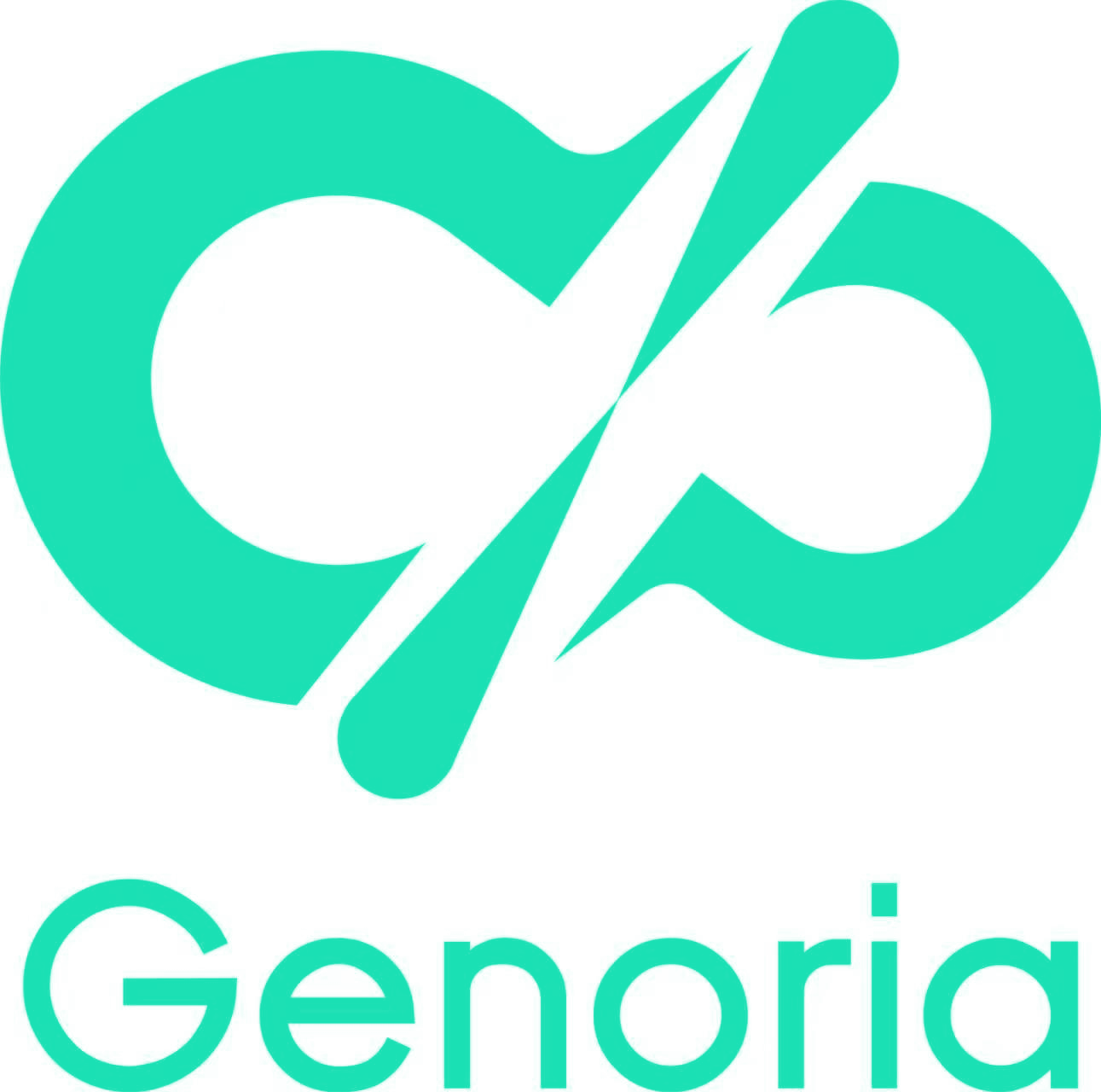}\hfill\begin{tabular}[b]{@{}r@{}}{\sffamily\Huge\bfseries\color{brand}Operon-S1}\\[0.3ex]{\sffamily\small\color{brandink}System 1 $\cdot$ fast decisions}\end{tabular}\par

\vspace{0.8ex}\noindent{\color{black}\rule{\linewidth}{1.6pt}}\par\vspace{1.4ex}

\begin{center}

{\sffamily\LARGE\bfseries\setstretch{1.05} Operon-S1:\\ a fast decision controller for long-horizon biomedical agents\par}\vspace{1.0ex}

{\color{black!60}\rule{0.9\linewidth}{0.4pt}}\par\vspace{1.2ex}

{\large\sffamily Genoria AI Team\par}\vspace{0.5ex}

{\normalsize\sffamily Yuejie Hou, Yueyuxiao Yang, Weiting Tang, Meng Yang\par}\vspace{0.15em}

{\small\color{black!65}Genoria AI\par}\vspace{0.3em}

{\small\color{black!65}29 October 2026\par}

\end{center}\vspace{0.8em}

\begin{abstract}Language-model agents now carry out long scientific work in biology, but their judgements about conventions, skills, errors, evidence and stopping are made inside the reasoning that does the work, where they cannot be measured, replaced or improved on their own. Here we present Operon-S1, a fast decision system, a System 1 for the agent, that separates judging from doing. A decision harness poses the agent's recurring judgements as closed questions at 13 decision points and acts only through fixed messages that never contain an answer, so any model that answers the questions can serve as the judge. Across seven life-science benchmarks, a local decision model took over these judgements in 0.6~s at the median for an in-loop request on CPU at no request cost, with no measurable change in the agent's score on five of six benchmarks, whereas the general language model that drives the agent, asked the same questions, accounted for 8.0\% to 35.2\% of task cost. Every decision is recorded with its consequence. Retraining on labels from task construction, verifiers or execution raised decision accuracy on every question in each of three trained versions, and a new version enters service only through a promotion gate. Task-level results in this study use the public checkpoint. The same closed decisions serve wet-laboratory protocol troubleshooting and the design of a germline target enrichment panel. Like the regulator of an operon, which switches genes on and off without making their products, Operon-S1 steers the agent without doing its work, and makes the agent's judgements a layer that can be measured, replaced and trained.\end{abstract}

\section*{Introduction}

Language-model agents now retrieve data, write and run analysis code, and translate scientific intent into executable laboratory procedures, in biology and biomedicine\cite{r1,r2,r3,r4,r5,r6}, where a general-purpose biomedical agent now carries out research tasks autonomously\cite{r2}, and in chemistry\cite{r7}. As these tasks lengthen, the limit on an agent shifts from what it can compute to how it judges its own work. A single analysis spans tens of steps, and along the way the agent meets questions that no tool answers: which convention a loosely worded task implies, which skill or file applies, what a failed command means, whether a value it has computed already answers the question, whether a slow computation should continue or yield its best result and, in laboratory work, which step of a protocol caused a failure. In such agents these judgements share one model's reasoning with the computation, so none of them is posed, measured or corrected on its own.

These judgements carry scientific weight. The independent-filtering threshold of a differential-expression analysis changes which genes are reported\cite{r8}, a skill whose convention differs from the task's can turn a correct analysis into a wrong answer, a misattributed protocol step sends the repeat experiment in the wrong direction, and a probe design judged by its design coverage alone can leave targets uncaptured. Once agents drive instruments, a misjudgement costs reagents and instrument time as well as compute. Reasoning agents can also favour extended internal reasoning over interacting with their environment, a failure termed overthinking\cite{r9}, and reasoning models keep deliberating after they have reached a correct answer\cite{r10}. In both cases more computation does not supply the missing judgement.

Cognitive science describes reasoning as the interplay of a fast, intuitive process and a slow, deliberate one\cite{r11,r12}, and it separates metacognitive monitoring, the assessment of one's own progress, from metacognitive control, the decision to continue, change strategy or stop\cite{r13}. Fast decision models, sometimes called System-One models, map a short state and a closed question to a distribution over a fixed set of options in a single forward pass\cite{r14,r15,r16}. Related approaches route queries between cheap and expensive models\cite{r17,r18,r19}, learn when to defer to a downstream decision maker\cite{r20} or score intermediate steps with learned verifiers\cite{r21,r22} and process reward models\cite{r23,r24}. Language models also act as judges of other models' outputs\cite{r25,r26} and as critics that flag errors in code\cite{r27}, and their verdicts are free text written by a model. Selective prediction, abstaining when uncertain, is the classical way to contain a fast model's errors\cite{r28,r29}, and conformal prediction gives such abstention distribution-free guarantees\cite{r30,r31}. In agent pipelines, System-One models have reduced calls to large models without loss of task success\cite{r32}, supported scientific decisions over prepared evidence\cite{r33}, scored the recurring decisions of search agents\cite{r34} and answered typed questions over clinical records\cite{r35}.

The same work also marks the limits of this approach. For a decision model fine-tuned on a browser agent's model-labelled decisions, held-out decision agreement rose from 19\% to 75\% without a corresponding gain in task success\cite{r36}, savings shrank once the cost of the decision model was counted\cite{r16}, and an open-weight local decision model changed 30\% of its answers when the option order was reversed\cite{r16}. In a cascade between a fast verifier and a large model, a matched random-escalation control showed that the savings came from how much was escalated rather than from the fast model's probabilities\cite{r37}. Scientific agents raise three further problems. The decisions that matter are scientific: conventions, analysis routes and whether the evidence suffices. A decision can sit tens of steps before the score it affects, because only the final answer is scored. And a fast model meant to improve as the agent works needs evidence of improvement from outside its own training loop.

An agent can hold its judgements in three ways, distinguished by two criteria: whether each judgement is posed on its own, and whether the model that answers it is trained and replaced against labels that no model produces. In the first, the judgements stay inside the reasoning of the model that does the work, and they change only when that model, its prompt or its skills change. In the second, judgements are lifted out to a fast model, one at a time\cite{r17,r20,r21} or several at once\cite{r34,r35}, and the model is judged by cost, by task success or against labels from people or language models. In the third, every recurring judgement of the run is posed as a closed question with a verifiable label, and the model that answers is replaced only under a gate. Operon-S1 takes the third route: it separates judging from doing.

Self-improving agents have refined their behaviour through verbal reflection on failures\cite{r38} and iterative self-feedback\cite{r39}, accumulated libraries of reusable skills\cite{r40}, rewritten their own code\cite{r41,r42} and evolved programs under automated evaluators\cite{r43}. In these systems a change is kept when it improves the signal that drives the search. A controller for scientific judgements needs a stricter test, because its labels are imbalanced and a candidate can score well by learning only the most frequent answer. Operon-S1 therefore keeps a new version only if it beats its parent and three baselines on labels that no model produces. Statistical adoption gates can control false adoptions in adaptive model search\cite{r44}, and no automated gate can in general both refuse every bad update and accept every good one\cite{r45}, so the gate here gives its guarantees only on a bounded, labelled split.

Here we introduce Operon-S1, a fast decision controller for long-horizon biomedical agents, paired with the ProtoPilot agent\cite{r1} (Fig.~1). As an external model, it takes over the two functions that metacognition research separates, monitoring and control\cite{r13}. The name refers to the operon\cite{r46}, in which a regulatory protein switches a cluster of genes on or off without making their products, as the controller regulates the agent's work without doing it, and ``S1'' to the fast process, System 1, of dual-process accounts of reasoning\cite{r11}. ProtoPilot coordinates protocol, computational and coding agents, verifies each stage against expert-grounded rubrics, and has generated standard operating procedures and instrument code that pass device-level validity checks across four laboratory automation platforms\cite{r1}.

Operon-S1 adds a decision layer with three parts to the agent, and ProtoPilot is the agent with which it is tested here. The decision harness is preset for each domain and triggers autonomously. It builds each state deterministically from the facts of the run, so the agent never assembles the context and every decision can be replayed, and it poses closed questions at 13 decision points from the start of a task to the submission of its answer (Table~1). The fast decision model answers each question with a probability for each option and acts through four fixed actions: brief, remind, end the tool loop or request one recheck. The local decision model, LAYA, runs on CPU. A hosted model, JEV, serves as a comparator, and a second comparator places a general language model in the same harness. LAYA is retrained in gated rounds on decisions from a data-synthesis pipeline. With the harness, the questions and the messages fixed, the system is designed to change only the decision model during retraining, so each change in behaviour can be attributed to it.

Each design choice follows from a property of the science. Because a judgement can decide a score without changing the computation, a message may name a convention, a file or a quantity to verify but never an answer value. Because conventions, evidence, error classes and answer formats can be checked by construction, verifiers or execution, each judgement is a closed question whose answer can be compared with such a label.

We find that this layer can be measured, replaced and traced at little cost. Six controllers, from rules to a general language model, ran on the same states, questions and messages, and in the 2,361 runs whose answers can be matched as text no injected message contained the gold answer. The local decision model answered single in-loop requests in under a second at the median and accounted for 0.8\% to 5.2\% of wall-clock time at no request cost. With its public checkpoint as the judge, the decision layer lowered cost on SeqQA by 5.8\% and agent steps on protocolqa2 by 10.3\%, with no measurable score change on five of six benchmarks. On seqqa2, every acting controller lowered the score, the local model least. Training on verifiable labels raised accuracy over the public checkpoint on every decision question in each of three trained versions, analysis family from 0.26 to 0.99 in the first. The gains carried over to the agent's own decision states on held-out benchmark items, where analysis-family accuracy rose from 0.056 to 1.000 and execution-labelled error-class accuracy from 0.174 to 0.408. The same closed decisions extend to laboratory-protocol troubleshooting, where the errors are of diagnosis rather than of routing, and to probe design, where a fall in panel coverage from 88\% to 16\% traces to three causes, each answered by one closed design action.

\begin{figure}[!tbp]
\centering
\includegraphics[width=\textwidth,height=0.72\textheight,keepaspectratio]{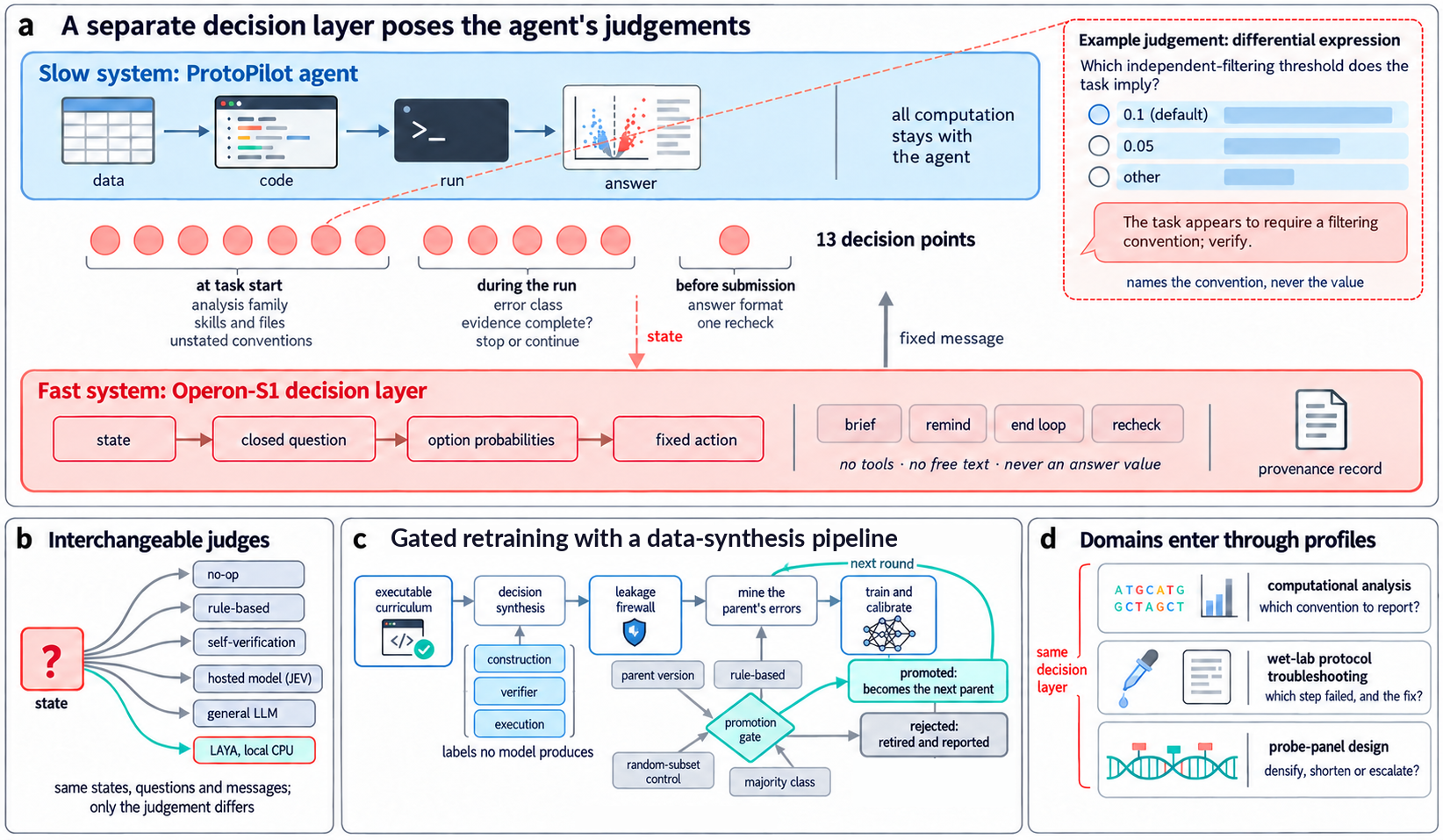}
\caption{\textbf{Operon-S1 separates judging from doing in a biomedical agent.} \textbf{a}, The slow system, the ProtoPilot agent, does all computation, from data and code to the answer. The decision harness poses the agent's recurring judgements at 13 decision points from the start of the task to its submission (Table~1). At each point the fast system, the decision layer, receives a state rendered from the facts of the run, answers a closed question with option probabilities and acts only through fixed messages, and each decision is kept in a provenance record. Right, a convention judgement in a differential-expression task: which independent-filtering threshold the task implies (Fig.~2b). The message names the convention and never the value; option lengths are illustrative. \textbf{b}, Six interchangeable controllers answer the same states, questions and messages and differ only in their judgements (Fig.~3; Extended Data Table~2). \textbf{c}, Gated retraining with a data-synthesis pipeline. Each round builds labelled decisions from an executable curriculum and trains a candidate, which enters service only through the promotion gate (Fig.~4a). Every trained version and its verdict are kept in an append-only registry (Table~3; Fig.~4). \textbf{d}, Domains enter through profiles of the same harness: computational analysis, wet-laboratory protocol troubleshooting (protocol profile) and probe-panel design (Fig.~5), each with one of its judgements. Schematic; no data are plotted.}
\label{fig:1}
\end{figure}

\begin{table}[!tbp]
\centering
\caption{\textbf{The 13 decision points of the decision harness, from the start of a task to the submission of its answer.} The harness layer triggers each point autonomously from the facts of the run, poses its closed questions and abstains when the controller is not confident. The last column gives the source of the training labels in round 1. Points marked – are recorded without training labels in round 1.}
\label{tab:1}
\footnotesize
\setlength{\tabcolsep}{4pt}
\begin{tabularx}{\textwidth}{ll>{\raggedright\arraybackslash}X>{\raggedright\arraybackslash}Xl}
\toprule
\textbf{Point} & \textbf{Judgement} & \textbf{Question (type)} & \textbf{Action} & \textbf{Round-1 label} \\
\midrule
S1 & analysis family & ten classes (choice) & scopes S6 & construction \\
S2 & skills & which skills may help, with a no-skill option (choice); relevance of the top three (yes/no) & brief: up to three skill names, never their text & construction \\
S3 & skill need & is a skill needed (yes/no) & recorded only & – \\
S4 & environment & which environment, when there is more than one (choice) & brief: environment path & – \\
S5 & files & is each input file relevant (yes/no) & brief: relevant files & – \\
S6 & conventions & convention per slot that the task does not state (choice) & brief: ``the task appears to require \ldots{}; verify'' & – \\
S7 & answer type & expected answer type (choice) & brief; scopes F1 & construction \\
F1 & evidence & does the evidence answer every part; is a step missing (yes/no ×2) & end run (file-answered tasks: remind) & – \\
P7 & long computation & continue / one more round / switch to cheaper / stop with best (score) & end run or remind & – \\
P3 & error triage & error class, ten classes (choice) & class-specific reminder & execution \\
P4 & convention in code & does the code conform to the convention (yes/no) & remind & – \\
P8 & workspace bounds & out of bounds (yes/no) & remind & – \\
D20 & answer check & does the answer conform (yes/no per slot) & one recheck naming what to verify (text-answered tasks only) & construction, verifier \\
\bottomrule
\end{tabularx}
\par\smallskip\begin{minipage}{\textwidth}\footnotesize On tasks with a protocol document, the protocol profile replaces S1 by a choice among ten wet-lab domains, adds the answer type protocol fix, adds a fixed reminder after a failed read of the protocol document, replaces D20 by three yes/no questions on specificity, consistency with the extracted protocol text and whether the change explains the symptom, and silences F1, P7, P3, P4 and P8 (Methods).\end{minipage}
\end{table}

\section*{Results}

\subsection*{Operon-S1 lifts the agent's judgements into a separate decision layer}

Operon-S1 separates judging from doing. The judgements of the agent go to a separate fast decision model through the decision harness, and all doing stays with the agent (Fig.~1a). The slow system is the ProtoPilot agent\cite{r1}, here driven by DeepSeek-V4.1-Flash at high reasoning effort, which plans, writes and runs code in a sandbox and composes the answer. The harness layer triggers autonomously during the run, reads its facts and poses closed questions. These cover the analysis family, the expected answer type, relevant files and candidate skills, and conventions that the task implies but does not state. They also ask whether the evidence already answers the task, whether a long computation should stop with its best result, which class an error belongs to, whether the code follows the required convention, and whether the answer meets its format and conventions (Table~1). In the terms of Nelson and Narens\cite{r13}, the answers to these questions are the controller's monitoring of the agent's progress, and the actions they trigger are its control.

The controller answers each question with a probability distribution over its options and acts only when it is confident, abstaining otherwise as in selective prediction\cite{r28}. Its actions are to brief the agent, remind it, end the tool loop with a final-answer instruction or request one recheck, each delivered through a message written in advance that may name a convention, a file or a quantity to verify. All controllers share this policy and these messages and differ only in the distributions they return, with one exception for the rule-based controller (Methods), so a comparison between controllers is a comparison between their judgements. Throughout, ``controller'' denotes any policy in the controller slot, including the no-op and rule-based controls, and ``fast decision model'' denotes the learned or hosted model behind it.

A no-op controller renders and records every state but always abstains, and leaves the agent's model requests unchanged (Methods). The effect of a controller $\phi$ on a task outcome $Y$ is the contrast
\begin{equation}
\Delta_\phi=\mathbb{E}\bigl[Y(\phi)\bigr]-\mathbb{E}\bigl[Y(\phi_{0})\bigr],
\label{eq:1}
\end{equation}
where $\phi_0$ is the no-op controller and the expectation is over tasks and runs; the controller's own latency is charged against each task's time limit. The evaluation has seven arms. The baseline is the original agent without the harness. The other six run the same harness with different controllers (Fig.~1b):

\begin{itemize}\setlength{\itemsep}{0pt}
\item{} the no-op control;
\item{} a rule-based controller that uses lexical matching, error-message patterns and budget fractions;
\item{} a self-verification control that requests a recheck whenever the answer check would pose a question;
\item{} JEV, a hosted System-One model\cite{r14}, used as a comparator;
\item{} LAYA, a local 322-million-parameter encoder\cite{r47} served on CPU, whose public checkpoint is version 0 of the trained controller;
\item{} a general-LLM controller, in which DeepSeek-V4.1-Flash answers the same closed questions in the same harness, with the same states, policy and messages.
\end{itemize}

The general-LLM arm measures what a fast decision model adds over asking a general language model the same questions. In every batch of the main table, JEV, LAYA and the general-LLM controller have equal time budgets for each request (Methods).

For every decision we keep a provenance record of the distribution, the action, whether the agent adopted it and the task outcome (Fig.~1a), which links a decision to a score assigned only to the final answer.

\subsection*{Judgements rather than computation decided most of the original agent's lost items and surplus time}

The judgements that the layer poses are where the original agent lost items and time. Most wrong answers of the original ProtoPilot agent on BixBench-Verified-50 arose from a judgement after the computation. The agent answered 12 of the 49 gradable items wrongly (Methods). On 5, the correct value was already in its output under another convention. On 1 it misread the direction of an effect, and on 2 it broke the required answer format. On 3 it lacked an offline resource, and on 1 none of its computed variants matched the reference. The first three groups, 8 of 12, turned on a choice made after the computation, where a controller can act. A further judgement, when to stop, costs mainly time, measured directly as the wall-clock time after the answer first appeared (Fig.~2a). For each task with a locatable answer we recorded the answer time $t^{\ast}$, the first moment at which the value the agent finally submitted appeared in its own tool output, and measured the post-answer fraction of wall-clock time
\begin{equation}
\omega=\frac{\sum_{i\in I}\bigl(T_i-t_i^{\ast}\bigr)}{\sum_{i\in I}T_i},
\label{eq:2}
\end{equation}
where $T_i$ is the wall-clock time of task $i$ and $I$ the set of tasks with a locatable answer (Methods). On the 15 BixBench-Verified-50 tasks with a locatable answer, $\omega=0.47$: 1,798 of 3,860~s elapsed after the answer was already in the agent's tool output. On SciAgentArena, where $t^{\ast}$ is taken as the first successful run of the solution script, 30\% of all wall-clock time on the benchmark followed $t^{\ast}$ (14,709 of 48,852~s), and on seqqa2 16\%; because tasks without a located $t^{\ast}$ add only to the denominator, these shares are lower bounds on $\omega$ (Methods). Because timed-out tasks score zero, continuing can also cost the whole item. On seqqa2, four of five timed-out codon-optimization tasks had produced a valid candidate 69–353~s into a run that then continued to the limit.

The convention errors share a pattern. On each of the 5 items, the reference value or the reference convention had already appeared in the agent's own output, but the agent submitted a value computed under another convention (Fig.~2b). In one case it ran a differential-expression analysis with the default independent-filtering threshold\cite{r48}, which reproduced the reference, and then reported the result of a different threshold. The difference between a right and a wrong answer was the choice between the default threshold of 0.1 and a threshold of 0.05.

Skill selection and error triage cost fewer items than conventions. Across 210 judged trajectories, loading a matching skill cost an item three times, because the skill text was wrong or its convention differed from the benchmark's, whereas skipping one cost an item once (Fig.~2c). On BixBench-Verified-50, SciAgentArena and BiomniBench-DA, the agent recovered unaided from 82–94\% of tool-error episodes, and the median error run lasted one call. Headroom therefore differs from one judgement to the next, which is why the layer poses, measures and trains each judgement on its own rather than repairing one failure class at a time.

The same analysis was applied to the baseline rounds of the formal evaluation, which used later skill releases and a higher reasoning effort (Methods; Extended Data Fig.~1). Of 121 wrong answers on the four benchmarks scored per item (SeqQA, seqqa2, BixBench-Verified-50 and SciAgentArena), a rule codebook assigned 9 to conventions, which it detects only where an answer matcher exists, 7 to stopping at the time limit and 1 to answer format. On 10 of 27 wrong answers with an automatic gold matcher, the correct value had already appeared in the agent's own output. On tasks with a locatable answer, the post-answer fraction $\omega$ was 0.33 on BixBench-Verified-50, 0.35 on SciAgentArena and 0.49 on seqqa2 (Extended Data Fig.~1).

\begin{figure}[!tbp]
\centering
\includegraphics[width=\textwidth,height=0.72\textheight,keepaspectratio]{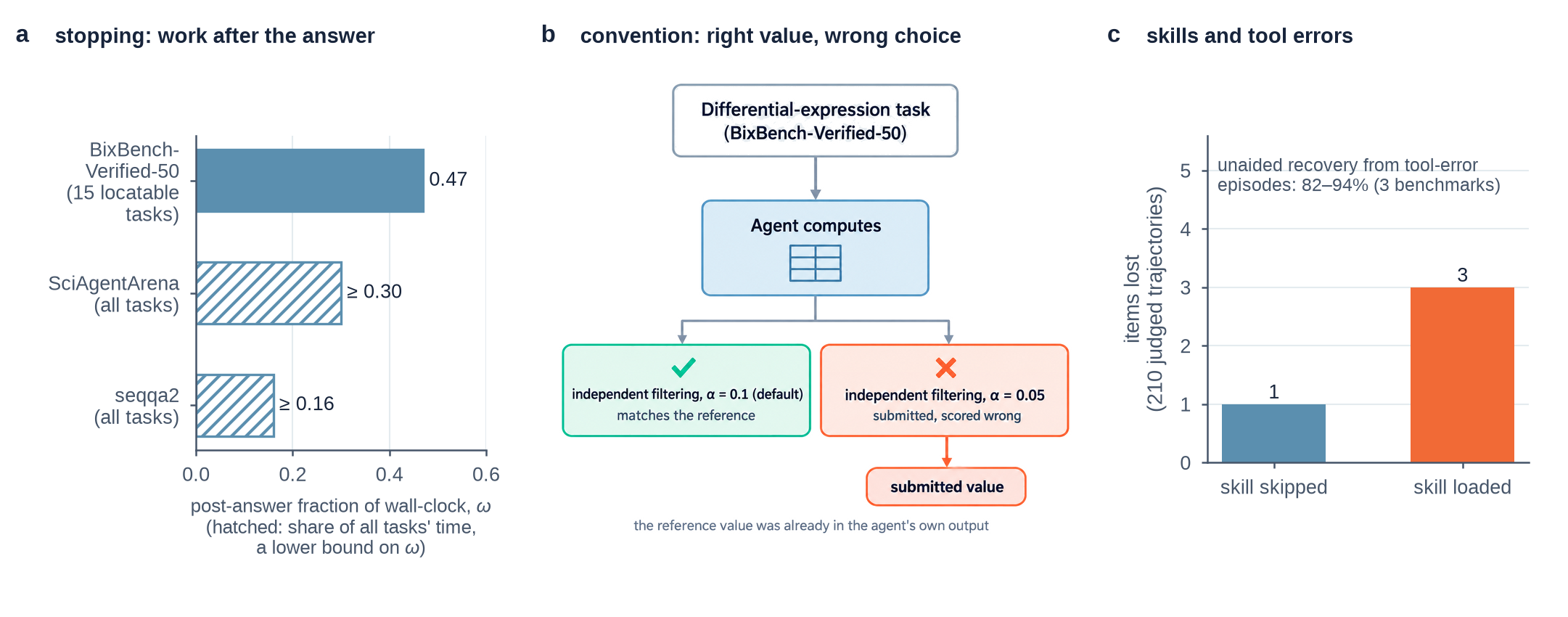}
\caption{\textbf{Judgements, not the computation, decided most of the original agent's lost items and time.} Trajectories of the original agent under two earlier skill releases. \textbf{a}, Post-answer time: $\omega$ (equation~(2)) on the 15 BixBench-Verified-50 tasks with a locatable answer, and the post-answer share of all wall-clock time on SciAgentArena and seqqa2, a lower bound on $\omega$. \textbf{b}, A convention mismatch in which the reference value appeared in the agent's own output. \textbf{c}, Items lost when a matching skill was skipped or loaded (210 judged trajectories); the annotation gives the range of unaided recovery from tool-error episodes on three benchmarks.}
\label{fig:2}
\end{figure}

\subsection*{The local decision model costs little beside the work it judges}

A separate judge pays off only if it costs little beside the work it judges and leaves the agent unchanged when it abstains. We evaluated on seven benchmarks spanning sequence reasoning, laboratory-protocol troubleshooting, end-to-end data analysis and multi-step omics studies, 1,417 tasks in total (Table~2): LAB-Bench SeqQA\cite{r49}, LAB-Bench 2 seqqa2 and protocolqa2\cite{r50}, BixBench-Verified-50\cite{r51,r52}, SciAgentArena\cite{r53}, BiomniBench-DA\cite{r54,r55} and the public split of OmicsBench\cite{r56}. Each benchmark ran under its official prompt, network policy and primary metric, with every task in a fresh sandbox. The original agent was scheduled for five rounds to estimate between-round variation, and Extended Data Table~2 gives the number of rounds that completed on each benchmark (Methods). Paired changes in time, steps and accuracy against the baseline are shown in Fig.~3a,b and Extended Data Table~2. The hypotheses are that the controller reduces cost at non-inferior accuracy (H1) and that its gains exceed those of an extra recheck (H2). H1 is judged on five benchmarks, 1,289 tasks: SeqQA, seqqa2, BixBench-Verified-50, SciAgentArena and BiomniBench-DA. H2 is judged on SeqQA and seqqa2. OmicsBench, on which the controllers are inert, is run in the baseline arm only. protocolqa2 was added to the evaluation after the hypotheses were written, so its results are reported as exploratory.

The no-op control checks that the harness itself leaves the agent unchanged. Against the baseline, the no-op arm changed the official score by −0.030 to +0.014 and wall-clock time by −6.9\% to +40.5\%, and every 95\% interval of its score, time and step changes included zero on the six benchmarks with paired tasks except the wall-clock change on BiomniBench-DA (+40.5\% [+18.0, +66.4]), where its batches ran partly at a different concurrency (Methods) (Extended Data Table~3). An audit of the 16,042 messages injected in 5,743 task runs found no message that contained the gold answer in any of the 2,361 runs whose answers can be matched as text. The remaining 3,382 runs have free-text, file or sequence-design answers that cannot be matched automatically.

Relative to the baseline, the decision layer with LAYA version 0 as judge lowered cost on SeqQA by 5.8\% [0.6, 10.8] at no request cost of its own; per task, JEV requests cost 0.19 to 0.48 thousandths of a dollar and general-LLM requests 2.1 to 9.8. On SciAgentArena the steps and cost of this arm changed by −6.8\% [−18.6, +3.8] and −4.7\% [−17.2, +7.6]. Its wall-clock change ranged from −14.0\% to +2.6\% across the hypothesis benchmarks, and its only score change with an interval excluding zero was −0.013 [−0.028, −0.001] on seqqa2 (Fig.~3a,b; Extended Data Table~2). JEV changed wall-clock time by −9.8\% to +6.9\% and steps by −6.2\% to +4.0\%, and lowered the seqqa2 score by 0.038 [0.018, 0.060]. The general-LLM controller changed wall-clock time by −5.2\% to +16.2\%, changed cost by −4.4\% to +61.5\%, and lowered the seqqa2 score by 0.048 [0.025, 0.072]. The rule-based controller reduced steps on SciAgentArena by 9.8\% [1.2, 19.3] and on seqqa2 by 5.5\% [0.9, 9.8], where its score fell by 0.018 [0.003, 0.035]. H1 requires an efficiency gain on at least four of the five hypothesis benchmarks and non-inferiority on every benchmark with a verdict (SciAgentArena, with one reference round, has none), and at the data cutoff it holds for no arm: the efficiency condition held on at most two benchmarks. Non-inferiority is limited by precision on two benchmarks. On BixBench-Verified-50 the score interval of the no-op control itself (−0.030 [−0.092, +0.023]) extended below the margin of 0.028, because 50 tasks in 33 clusters give an interval wider than the margin, and on SeqQA the margin is 0.0023, because two baseline rounds that agree closely give a between-round standard deviation of 0.001 (Methods).

On protocolqa2, an exploratory benchmark, LAYA version 0 and the general-LLM controller reduced steps by 10.3\% [4.9, 15.7] and 10.9\% [4.8, 16.8] with score changes of −0.008 and 0.000, and JEV changed the score by +0.032 [−0.024, +0.088]. On BiomniBench-DA, with all 50 tasks paired, JEV, LAYA version 0 and the general-LLM controller reduced steps by 16.6\% [9.1, 23.2], 13.9\% [5.9, 21.2] and 16.1\% [9.5, 22.3] with no score change outside its interval, and the changes of JEV and LAYA are for reference only, because their batches ran at a different concurrency (Methods).

These figures include the controller's own requests, latency and cost. In offline replay of 8,265 gate rows, the local controller answered in 266~ms at the median and 581~ms at the 95th percentile on CPU. During the evaluation, the median error-triage request (P3) took 576~ms for LAYA, 222~ms for JEV and 1,462~ms for the general-LLM controller, against 8 to 34~s for a slow-system step across benchmarks. The brief issues two sequential requests over the longest states. It took 7.3~s at the median on the local model, against a median of 164~s for a whole SeqQA task. The brief carries 7 to 31 questions and its requests queue on four shared replicas, whereas single in-loop requests took 0.5 to 0.8~s at the median (workspace bounds 531~ms, error triage 576~ms, answer check 667~ms, convention in code 811~ms, long computation 839~ms). Efficiency is reported per benchmark. Across benchmarks, the controller accounted for 0.3\% to 1.4\% of wall-clock time and 0.4\% to 2.2\% of cost per task for JEV, and for 1.9\% to 16.9\% and 8.0\% to 35.2\% for the general-LLM controller (Extended Data Fig.~2). LAYA version 0 accounted for 0.8\% to 5.2\% of wall-clock time and no request cost.

\begin{figure}[!tbp]
\centering
\includegraphics[width=\textwidth,height=0.72\textheight,keepaspectratio]{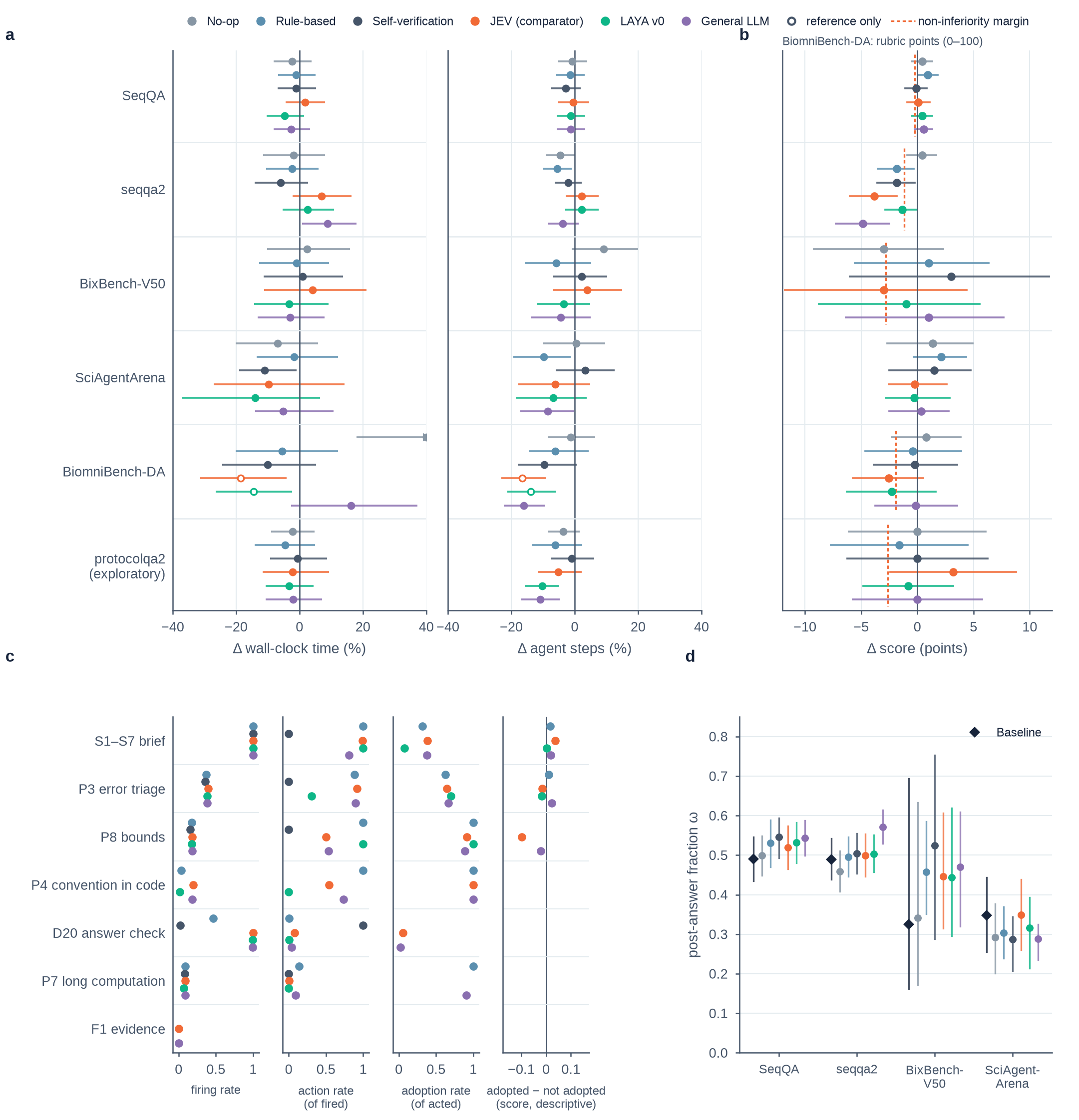}
\caption{\textbf{Efficiency, accuracy and decision provenance of six controllers in one harness.} \textbf{a}, Paired relative change in wall-clock time and agent steps against the baseline (ratio of paired sums minus one) per controller and benchmark. Tasks: SeqQA 600, seqqa2 400, BixBench-Verified-50 50, SciAgentArena 189, BiomniBench-DA 50 (common set at the data cutoff) and protocolqa2 125. Hollow markers, BiomniBench-DA changes of JEV and LAYA, shown for reference (Methods). protocolqa2 is exploratory. \textbf{b}, Paired change in the official score in points (accuracy × 100; BiomniBench-DA rubric points, 0–100), with the non-inferiority margin (dashed). SciAgentArena has a single reference round and no margin. Arrows mark intervals beyond the axis. \textbf{c}, For each decision point, pooled over the six benchmarks: firing rate per task, action rate as a share of firings, adoption rate as a share of actions whose adoption is defined, and the outcome difference between adopted and non-adopted cases, which is descriptive because adoption is not randomized. The difference is weighted by tasks over benchmarks scored 0–1 with at least five cases in each group and is drawn only when at least ten cases are pooled. \textbf{d}, Post-answer fraction $\omega$ (equation~(2)) on tasks with a locatable answer, per arm and benchmark. Error bars, 95\% cluster-bootstrap intervals.}
\label{fig:3}
\end{figure}

\begin{table}[!tbp]
\centering
\caption{\textbf{Benchmarks and settings.} Per-task time limits are twice the official or default limit (Methods).}
\label{tab:2}
\footnotesize
\setlength{\tabcolsep}{4pt}
\begin{tabularx}{\textwidth}{lr>{\raggedright\arraybackslash}Xl>{\raggedright\arraybackslash}X>{\raggedright\arraybackslash}X}
\toprule
\textbf{Benchmark} & \textbf{Tasks} & \textbf{Metric} & \textbf{Network} & \textbf{Judge} & \textbf{Per-task limits} \\
\midrule
LAB-Bench SeqQA & 600 & accuracy & database allowlist & none & 1 CPU, 2~GiB, 1,200~s \\
LAB-Bench 2 seqqa2 & 400 & accuracy (official validators) & database allowlist & none & 1 CPU, 2~GiB, 1,200~s, last 60~s reserved \\
LAB-Bench 2 protocolqa2 & 125 & accuracy (official judge prompt) & offline & LLM (substitute model) & 2 CPU, 4~GiB, 2,400~s \\
BixBench-Verified-50 & 50 & open-answer accuracy & database allowlist & LLM for range answers (substitute model) & 2 CPU, 16~GiB, 3,600~s \\
SciAgentArena & 189 & mean correctness & offline & none & 4 CPU, 8~GiB, 3,600~s \\
BiomniBench-DA & 50 & rubric score, 0–100 & web via logged proxy & LLM rubric (substitute model) & 2 CPU, 16~GiB, 7,200~s \\
OmicsBench (public) & 3 & mean node score & web via logged proxy & official & 4 CPU, 12~GiB, 43,200~s \\
\bottomrule
\end{tabularx}
\end{table}

\subsection*{Provenance records separate whether a judgement is right, whether the agent acts on it and whether the task succeeds}

Monitoring and control are measured separately. Monitoring quality is the agreement of each judgement with its label, by per-question accuracy and calibration (Extended Data Fig.~3; Extended Data Table~6). On benchmark states with execution- or benchmark-derived labels (error class, analysis family and answer type), per-question accuracy ranged from 0.865 to 1.000 for JEV and from 0.855 to 0.998 for the general-LLM controller, with expected calibration errors of 0.013 to 0.039 and 0.033 to 0.045. Version 0 of LAYA, the public checkpoint used zero-shot, was poorly matched to these states: 0.164 on error class (expected calibration error 0.360), 0.087 on analysis family and 0.663 on answer type, against 0.994 on answer format, where JEV scored 0.680 and the general-LLM controller 0.903 (Extended Data Table~6). Round 1 trains these questions on curriculum states. At the action threshold of 0.6, on error triage, JEV answered 91.2\% of questions with an accuracy of 0.911 on those answered, and the general-LLM controller answered 92.9\% with an accuracy of 0.882 (Extended Data Fig.~3). Control is measured by the contrast of equation~\eqref{eq:1}, together with the post-answer fraction $\omega$ of each arm (Fig.~3d). In the baseline $\omega$ was 0.49 on SeqQA, 0.49 on seqqa2, 0.33 on BixBench-Verified-50 and 0.35 on SciAgentArena. Three arms changed it with an interval excluding zero: the general-LLM controller raised it on seqqa2 by 0.081 [0.036, 0.127], and the rule-based and self-verification controls lowered it on SciAgentArena by 0.044 [0.001, 0.092] and 0.061 [0.016, 0.123].

The self-verification control isolates the value of looking again before the answer is submitted. It uses the same states, triggers and messages as the learned controllers but requests a recheck whenever the answer check would pose a question. On SeqQA and seqqa2, JEV differed from the self-verification control by +0.002 [−0.010, +0.013] and −0.020 [−0.043, +0.003] in accuracy, at 6.5\% and 10.2\% higher cost. H2 is therefore not supported for JEV. LAYA version 0 and the general-LLM controller differed from the self-verification control by +0.005 and +0.005, and by +0.007 and −0.030, in accuracy on SeqQA and seqqa2, with every interval including zero, an exploratory comparison.

Against the baseline, on SeqQA, seqqa2, BixBench-Verified-50 and SciAgentArena together, LAYA version 0 turned 13 wrong answers right and 21 right answers wrong. JEV did so for 15 and 38 answers, and the general-LLM controller for 16 and 40 (exact McNemar test\cite{r57} with Holm correction across benchmarks; Extended Data Table~4). Among the McNemar tests, after Holm correction the seqqa2 losses of the rule-based controller (0 gained, 12 lost, P = 0.002), the self-verification control (2 gained, 14 lost, P = 0.021), JEV (1 gained, 21 lost, P < 0.001), LAYA version 0 (0 gained, 10 lost, P = 0.010) and the general-LLM controller (1 gained, 25 lost, P < 0.001) remained significant, and no other test did. The seqqa2 losses of JEV and the general-LLM controller were concentrated in the synonymous-mutation and molecular-weight families, and changes fell on items of medium and high baseline difficulty (Extended Data Fig.~4). Because the loss is visible decision by decision, it can be located. On the 23 development items on which JEV lost answers that the baseline got right, the agent loaded the convention skill in 13 runs under JEV against 16 in the baseline.

The provenance records locate where effects arise (Fig.~3c). For each decision point we report how often it fired, how often the controller acted, how often the agent adopted the action and the outcome difference between adopted and non-adopted cases, a descriptive quantity because adoption is not randomized. The brief fired on every task, and the agent loaded a suggested skill after 38\% of briefs under JEV and 7\% under LAYA version 0. The answer check requested a recheck on 8.0\% of its firings under JEV. The evidence check fired on at most one task per benchmark, and the long-computation check acted on 33\% of its firings under the rule-based controller and 14\% under the general-LLM controller on SciAgentArena. Workspace-bounds and convention-in-code reminders were adopted in 82\% to 100\% of cases (Fig.~3c).

\subsection*{Gated retraining couples a data-synthesis pipeline with a promotion gate}

LAYA is retrained in gated rounds (Fig.~1c, Fig.~4a). Each round runs the data-synthesis pipeline, which turns an executable curriculum into decisions with verifiable labels, mines hard examples from the parent version's errors, trains a candidate, fits a temperature-scaling calibration\cite{r58} for it, and tests it against a promotion gate. A promoted version controls the agent and is the parent whose errors the next round mines, so each round starts from the last promoted version.

Round 1 draws its decisions from a synthetic curriculum of 1,517 executable bioinformatics tasks in 25 families, each with two independent implementations, metamorphic checks\cite{r59} and a verifier that accepts its reference answer and rejects constructed negatives. Four generators turn these tasks into decisions with the controller's own state renderer and questions, so that training states have the form of deployment states (Extended Data Table~1). Labels come only from construction, a verifier or code execution. This follows programmatic supervision\cite{r60} but needs no label model, because these labels are exact. A fault example is kept only if running it fails. Round 1 builds every decision from templates. A leakage firewall admits only rows with these label sources, no lineage through the comparator and no overlap with the benchmarks' task text\cite{r61,r62} (Methods).

A loss-weighted variant of hard-example mining\cite{r63} samples rows within each question with a weight that increases with the parent version's loss and doubles for confident errors. Each candidate is paired with a random-subset control trained on the same number of rows per question, which gives the gate a baseline of equal training budget. Let $\mathcal{Q}^{\ast}$ be the questions with headroom, on which the rule-based controller scores at most 0.95. A candidate $\theta'$ replaces its parent $\theta$ only if $\mathcal{Q}^{\ast}\neq\varnothing$ and, for every $q\in\mathcal{Q}^{\ast}$,
\begin{equation}
\begin{gathered}
Q_{0.025}\bigl[\widehat{\Delta}_q^{\ast}\bigr]>0,\qquad
\mathrm{ECE}_q(\theta')\le0.10,\\
\mathrm{acc}_q(\theta')>\max\bigl\{\mathrm{acc}_q(\mathrm{rule}),\,\mathrm{acc}_q(\mathrm{random}),\,\mathrm{acc}_q(\mathrm{majority})\bigr\},
\end{gathered}
\label{eq:3}
\end{equation}
where $\widehat{\Delta}_q=\mathrm{acc}_q(\theta')-\mathrm{acc}_q(\theta)$ is computed on identical states and $Q_{0.025}$ is the lower 2.5\% quantile of its cluster-bootstrap replicates $\widehat{\Delta}_q^{\ast}$ over entities. ECE is the expected calibration error of the uncalibrated candidate, and ``random'' is the random-subset control. The full rule (equation~\eqref{eq:14}) adds a minimum of 200 labelled rows per question, a regression limit of one point and a refusal of candidates whose answers collapse onto one option. Release gates have been proposed for self-improving agent runtimes as an engineering safeguard\cite{r64}, and here the gate is also part of the scientific claim. The gate releases a version only if it beats its parent, the rule-based controller, the majority class and a random-subset control trained on the same number of rows, and every trained version, its data, checkpoint and verdict are kept in an append-only registry. When a round learns from execution-labelled benchmark states, each benchmark task is answered only by a version whose lineage never trained on its group, an out-of-fold scheme analogous to cross-fitting\cite{r65} and stacked generalization\cite{r66}.

The first round shows why the gate needs a majority-class baseline. Decision labels are imbalanced: on the gate split, 92\% of analysis-family labels belong to a single class and 87\% of skill-relevance labels are negative (Extended Data Table~1). Version 0, the public checkpoint used zero-shot, scored 0.26 on analysis family against 0.92 for the majority class and 0.79 for the rule-based controller, and 0.02 on skill choice against 0.69 and 0.68 (Fig.~4b). The skill-choice score is also below the chance level of about 1/16 for 16 options. On answer type and answer format the checkpoint scored 0.40 and 0.44, against majority-class rates of 0.49 and 0.50, and only on error class, with ten classes, did it exceed the majority class (0.43 against 0.08; rule-based controller 0.85). The rule-based controller is itself below the majority class on four of the six questions, so it cannot expose a candidate that learned only the label prior. Under such imbalance accuracy misleads, the accuracy paradox\cite{r67}: a candidate that always gave the most frequent training label would score 0.92 on analysis family and 0.87 on skill relevance, tens of points above version 0, and of the baselines in equation~\eqref{eq:3} only the majority class is certain to refuse it. Balanced accuracy\cite{r68} corrects for the imbalance, but the gate's verdict rests on accuracy against the three baselines.

A judgement enters retraining as soon as it can be labelled without a model. Round 1 therefore trains the judgements that construction, verifiers or execution label on the curriculum: analysis family, answer type, skill choice and relevance, answer format and error class.

\begin{figure}[!tbp]
\centering
\includegraphics[width=\textwidth,height=0.72\textheight,keepaspectratio]{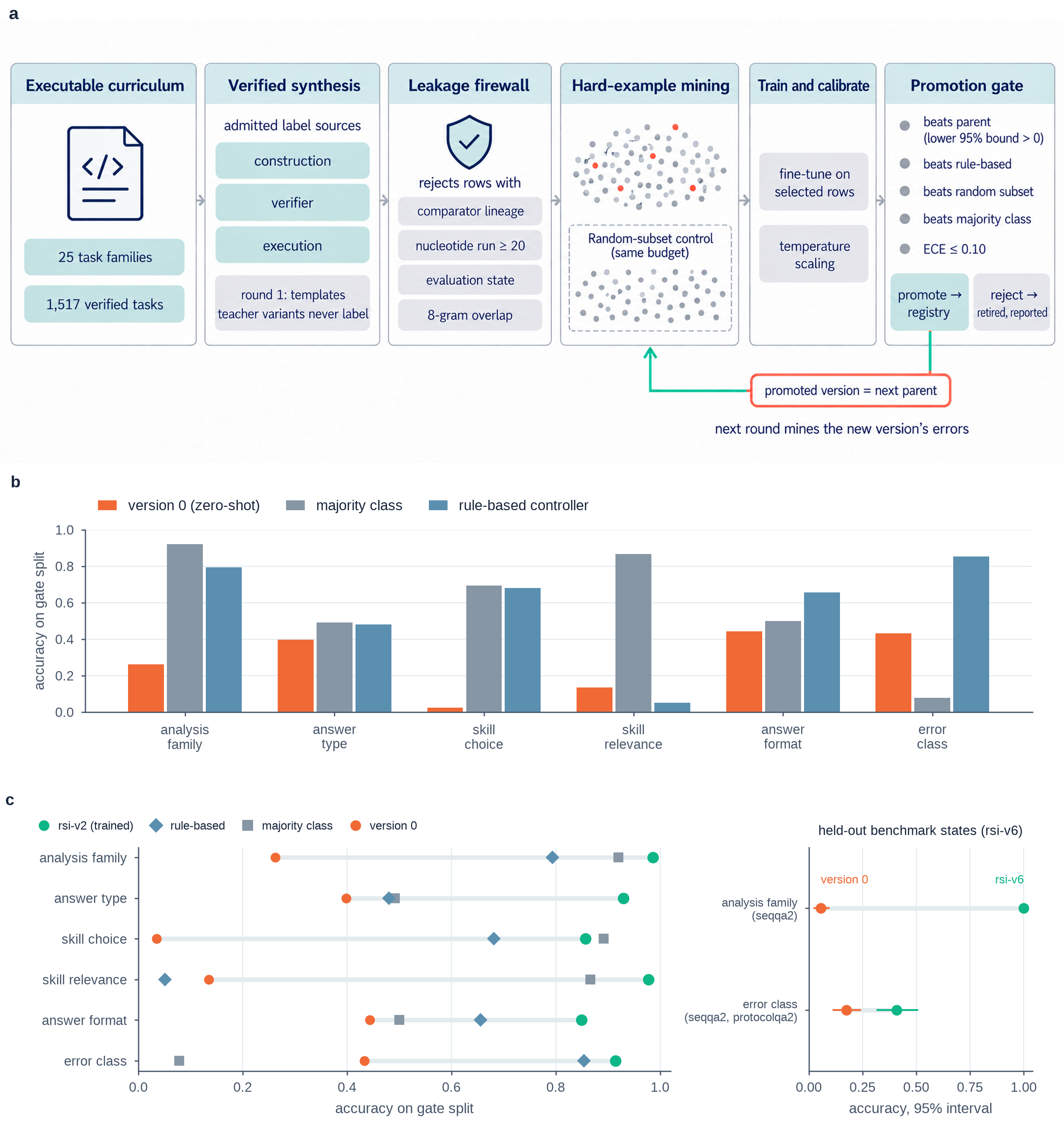}
\caption{\textbf{Gated retraining of the fast decision model.} \textbf{a}, One round of retraining: the data-synthesis pipeline (verified synthesis from an executable curriculum), leakage firewall, hard-example mining with a random-subset control of equal budget, training and calibration, and the promotion gate, whose conditions are given in equations~(3) and (14); promoted versions enter the registry and the next round mines their errors. \textbf{b}, Gate-split accuracy per question of the zero-shot public checkpoint (version 0), the majority class (the most frequent training label) and the rule-based controller. Skill choice is scored against the single constructed label, and skill relevance pools all skills. \textbf{c}, Training raised accuracy on every decision question. Left, gate-split accuracy per question of version 0 and the first trained version, rsi-v2, with the majority class and the rule-based controller, with skill choice scored against each state's set of valid skills, as the gate does. Right, accuracy of version 0 and rsi-v6 on the agent's own decision states from held-out benchmark items (analysis family on seqqa2; error class labelled by execution, pooled over the held-out halves of seqqa2 and protocolqa2), with paired 95\% cluster-bootstrap intervals.}
\label{fig:4}
\end{figure}

\subsection*{Training raised accuracy on every decision question, and the gains carried over to the agent's own states}

The first trained version, rsi-v2, was trained on 2,500 rows of curriculum decisions and raised accuracy over version 0 on all six questions of the gate split, with lower bounds above zero. Under the gate's scoring, in which a skill choice is correct if it is among the state's valid skills, analysis family rose from 0.26 to 0.99, skill choice from 0.04 to 0.86, skill relevance from 0.13 to 0.98, error class from 0.43 to 0.91, answer type from 0.40 to 0.93 and answer format from 0.44 to 0.85 (Fig.~4c). Its calibration error was at most 0.08 on every question.

Further rounds raised accuracy in the same way (Table~3). rsi-v5, trained on synthetic protocol decisions, raised accuracy over version 0 on all four questions of its gate split, with lower bounds above zero: analysis family from 0.11 to 0.53, answer type from 0.24 to 1.00, skill choice from 0.04 to 0.46 and skill relevance from 0.53 to 0.68. rsi-v6 was trained on a curriculum shaped by the agent's own trajectories: its training mix, the number of rows per decision type, was set by the decision errors that the controllers made on the development halves of seqqa2 and protocolqa2 in the main run. Its rows came from the curriculum generators and the verified synthetic protocols with labels by construction, so no development item became a training row, and each arm was trained on 1,200 rows, 1,000 drawn for it and 200 shared replay rows (Methods). On a held-out gate split of 250 generated decisions per question, it raised accuracy over version 0 on all four questions, with lower bounds above zero: analysis family by 0.868 [0.819, 0.912], answer type by 0.656 [0.605, 0.707], skill choice by 0.844 [0.799, 0.885] and skill relevance by 0.452 [0.365, 0.532]. Because decision labels are imbalanced, we also report balanced accuracy, the mean recall over classes: every trained version exceeded the majority class on every question (24 of 24 question–split pairs, +0.12 to +0.81).

These gains carry over from generated decisions to the agent's own states. Because only the decision model changes, a trained version can be scored on the agent's own decision states from the held-out halves, whose items training never drew on (Methods). On these states, rsi-v6 raised analysis-family accuracy on seqqa2 from 0.056 to 1.000 (+0.944 [+0.908, +0.974]; calibration error from 0.449 to 0.015; one label value) and execution-labelled error-class accuracy from 0.174 to 0.408 (item-level paired difference +0.186 [+0.045, +0.329]), the two judgements on which version 0 was weakest on benchmark states (Extended Data Table~6). Training also made the decision model robust to option order: re-asked with the options in another order, the public checkpoint changed 37\% [34\%, 40\%] of its answers, in line with the sensitivity reported for open-weight decision models\cite{r16}, and rsi-v6 changed 6\% [4\%, 8\%] (750 choice questions on gate states). On the same states, the version trained on rows selected by the parent's errors changed about half as many answers as a version trained on a random subset of the same size (5.6\% [4.2\%, 7.5\%] against 11.3\% [9.3\%, 13.8\%]), so error selection made the judgements more stable. The quantity that training targets and the gate certifies is decision accuracy. Whether a correct decision changes a task's score depends further on whether the agent acts on it and on the headroom of the judgement (Fig.~2; Fig.~3c), which the main run measures with version 0 as its LAYA arm.

\begin{table}[!tbp]
\centering
\caption{\textbf{Versions of the decision model.} Every version was judged under gate-rules@2 (equations~(3) and (14)). Δ v0, gain over version 0 on the gate split with paired 95\% cluster-bootstrap intervals. Held-out states, accuracy on the agent's own decision states in held-out benchmark runs.}
\label{tab:3}
\footnotesize
\setlength{\tabcolsep}{4pt}
\begin{tabularx}{\textwidth}{l>{\raggedright\arraybackslash}X>{\raggedright\arraybackslash}X>{\raggedright\arraybackslash}X>{\raggedright\arraybackslash}X}
\toprule
\textbf{Version} & \textbf{Training data} & \textbf{Gate split} & \textbf{Δ v0} & \textbf{Held-out states} \\
\midrule
Version 0 & public checkpoint, zero-shot & – & – & seqqa2 family 0.056; error class 0.174 \\
rsi-v2 & 2,500 rows of curriculum decisions & 6 questions & positive on all 6 (family 0.26 → 0.99) & – \\
rsi-v5 & synthetic protocol decisions, labels by construction & 4 questions & positive on all 4 (answer type 0.24 → 1.00) & – \\
rsi-v6 & 1,000 new synthetic rows and 200 replay rows per arm, mix set by controllers' errors on the development halves & 250 rows per question, 4 questions & +0.452 to +0.868, lower bounds above 0 & seqqa2 family 1.000; error class 0.408 \\
\bottomrule
\end{tabularx}
\end{table}

\subsection*{The seqqa2 baseline gain comes from a convention skill layer, measured apart from the decision layer}

Skills and the decision layer can both shape a judgement, such as which convention to report, so we measured their contributions separately. SeqQA, seqqa2, BixBench-Verified-50 and protocolqa2 ran with skill release r04b. The other benchmarks ran with release r03. r04b adds a general sequence-computation layer and a separately named convention layer for LAB-Bench to r03, together with a patch to the protocol-audit skill. The conventions in that layer were taken from the published LAB-Bench 2 scoring code. Skills were developed only on the development half of seqqa2, split by a hash of the item identifier, and the held-out half was tested once per release. On those 196 held-out items, at high reasoning effort, accuracy was 90.8\% with the convention layer and 77.4\% without it (151 of 195 gradable items). The held-out comparison for r03 comes from a medium-effort run (76.0\%, 149 of 196). In a paired ablation under the same conditions, high and medium effort differed within noise (Methods). On a random set of 60 SeqQA items, accuracy was 60 of 60 with the convention layer and 58 of 60 without it. On BixBench-Verified-50, r04b and r03 both answered 43 of 50 items correctly. The gain on seqqa2 therefore comes from the convention layer, and the seqqa2 baseline of the main run is reported with that provenance.

Between the earlier round and the main run, the JEV effect on seqqa2 changed from +0.008 to −0.041, a change of −0.048 [−0.081, −0.016], most of it in the synonymous-mutation family, and on SeqQA from +0.020 to +0.001 (Extended Data Fig.~5). On protocolqa2, which ran under both releases, r04b changed the baseline score by +0.030 [−0.013, +0.072] relative to r03, and the JEV effect changed from −0.018 under r03 to +0.031 under r04b, a change of +0.048 [−0.031, +0.127]. The r04b baseline took 91\% longer per task, and the two series also differ in run time and in the frozen code tree, so this comparison does not isolate the skill effect. Between the earlier round and the main run, both the reasoning effort and the skill release changed. That comparison is reported as their joint effect.

\subsection*{A domain profile extends the same harness to laboratory-protocol troubleshooting}

Laboratory-protocol troubleshooting needs different judgements from computational analysis. In protocolqa2 the agent reads a protocol and a reported failure, and it must name the step that caused the failure and the change that fixes it. The judgements are of a different kind. On the development half (62 items, about 17 graded runs each across the baseline rounds and controller arms), 780 of 1,060 graded answers were correct, and 14 items held 191 of the 280 wrong answers. Most wrong answers blamed a salient but standard parameter or step, located the fault at the wrong step, or used the protocol's controls only in part; others were distracted by a second oddity, restored the wrong original value, missed an absent or swapped reagent or trusted an altered reference value in the document. These are diagnosis errors rather than routing errors. The default harness was built for computational analyses: its analysis families and answer types do not describe a protocol fix, and its skill shortlist ranked a general protocol-repository skill ahead of the matching wet-lab skills.

We therefore built a protocol profile of the harness. It applies only to tasks whose workspace contains a protocol document, and on other tasks the harness is unchanged. The profile ships with the skill release as a catalog of ten wet-lab domains and their skills, and it adds the answer type ``protocol fix''. For the answer it asks three closed questions: whether the answer names one step, reagent or parameter with a concrete change, whether the cited step matches the protocol text the agent extracted, and whether the change would explain and fix the reported symptom. A failed attempt to read the protocol document triggers a fixed reminder that lists the text-extraction tools available in the sandbox. As with every message, no catalog entry, template or reminder contains an answer or item-specific content.

The profile was designed after the protocolqa2 results of the main run were seen, and it was tuned on a development half of protocolqa2. It is therefore exploratory and outside the hypotheses. On the 62 held-out items, tested once and held out from tuning, although their main-run outcomes under r03 had been seen, JEV with the protocol profile answered 0.823 correctly, against 0.758 for the baseline under the same development skill release (5 items gained, 1 lost, sign test P = 0.22). In the main run, three rounds of JEV and of the general-LLM controller with the protocol profile changed the score by +0.011 [−0.023, +0.045] and +0.035 [−0.002, +0.073] relative to the baseline, at 11.7\% and 15.0\% less wall-clock time, and by −0.021 and +0.035 relative to the same controllers under the default harness. With the local decision model as judge, the protocol profile raised the score by +0.060 [+0.014, +0.108] relative to the baseline at 7.9\% fewer steps, and by +0.068 [+0.012, +0.128] relative to the same model under the default harness (two rounds; Extended Data Table~5).

On synthetic protocol probes, LAYA version 0 chose wet-lab domains correctly, but its answers to the three answer questions ran in the wrong direction. It rated a generic answer as more specific than a specific one, so the profile has LAYA abstain on those questions. To train them we built laya-proto from synthetic protocol data with labels by construction. Claude Opus 5.5 wrote 128 of the 154 protocols from domain and topic prompts written by the authors, without benchmark text. DeepSeek-V4-Pro, at maximum reasoning effort, wrote the remaining 26 from the same prompts. Each protocol came with four to six single-substring faults, each with an observation and its correction. Two verifiers checked every fault: a structural check, and an independent diff check that the faulted protocol differs from the clean one at exactly one line, at the stated step. Labels are assigned by construction from the teacher-written fault and its correction, which the diff verifier checks, and no model assigns a label. The domain and skill come from the generating topic, and the answer type from the scenario template. The three answer questions are labelled from candidate answers built to be correct, to cite a wrong step, to misquote the right step, or to stay generic. Splits are by topic group, so no protocol or topic crosses splits. A bench firewall, whose corpus was never sent to the teacher, rejected any protocol that shared a title or a word 8-gram with the protocolqa2 questions or reference answers (0 of 154), and any row whose state shared a word 8-gram with those texts or with the staged protocol files (535 training, 167 calibration and 247 gate rows). The gate judges domain, answer type, skill choice and skill relevance, and the three answer questions are kept with their held-out protocols. laya-proto was judged by the gate as rsi-v5 (above).

\subsection*{Probe design poses the same closed decisions over measured state}

Probe design separates measuring from judging in the same way as analysis. A germline target enrichment panel uses hybridization probes to enrich selected genomic regions before sequencing. Its probes are useful only if they bind their targets and nothing else, remain thermodynamically stable without folding on themselves\cite{r69}, and cover each target evenly enough to be read\cite{r70,r71,r72}. Germline panels test all of these conditions, because paralogous gene families contain near-identical exons\cite{r73}, exons can be GC-extreme or shorter than a probe, and population alleles introduce mismatches at binding sites. In Operon-S1, probe design is divided along the same lines (Fig.~5a). Tools measure every sequence property: tiling with MrBait\cite{r74}, genome-wide alignment with BWA-MEM\cite{r75} and BLAST\cite{r76}, nearest-neighbour melting temperature, hairpin and self-dimer checks, and coverage. A slow agent plans and handles exceptions over a shared panel state, in a pre-synthesis design loop and a post-experiment redesign loop. What remains are closed choices over the measured state among eight design actions: keep, densify, relax the GC or melting-temperature window, lengthen, shorten, tighten the specificity screen, escalate to another assay, or stop. The decision layer poses these choices as closed questions over the measured state.

The first of these decisions, per-probe quality control, was measured directly. On 150 held-out keep-or-redesign decisions (75 of each label; 25 target groups held out from training; labels from the genome-wide specificity screen), the hosted decision model answered 0.913 [0.857, 0.949] correctly, against 0.500 for the public LAYA checkpoint, the chance level of this balanced set (Methods). One rule over visible features, fitted to the labels, reproduces all 3,859 keep-or-redesign labels in-sample, so this decision serves to check the layer against a known answer. The decisions that remain open are the method and parameters for a region, the attribution of low coverage of unknown cause, the protocol for GC-extreme regions and the choice among candidate panels.

On a 123-target, 31,855-bp germline panel on GRCh37\cite{r77}, a default design of 549 probes (90~nt, 45-bp step) reached a mean per-target coverage of 88.4\% (Fig.~5b). After genome-wide specificity screening, 207 probes passed, 24 targets kept at least one passing probe and 16 were fully covered, and mean per-target coverage by passing probes fell to 15.8\% (Methods). Because 82 of the 123 targets belong to the C4A/C4B pair, the other 41 targets are also reported separately: their mean coverage fell from 71.2\% to 47.5\%, and all 16 fully covered targets are among them. A denser design (70~nt, 35-bp step) covered every target base before screening, yet only 21 of 123 targets were fully covered after it (mean 20.4\%; 61.2\% outside C4A/C4B), and a targeted rescreen of the gaps raised fully covered targets from 16 to 19. Design coverage, post-screen coverage and capture are therefore different quantities, and the design summary alone overstates what a panel covers after screening.

The remaining gaps had three causes, each calling for a different action (Fig.~5c): sequence shared with another locus, exons shorter than a probe and GC-rich intervals. In the complement C4A/C4B pair, 35 of 41 exon pairs are identical in sequence. Shared probes gave a mean per-target design coverage of 96.9\% over the pair's targets, and none passed screening, because no probe can tell the two genes apart. At four boundary targets in two high-identity paralog pairs, none of 1,342 candidate units yielded a gene-specific passing probe, whereas 571 of 839 passed at a more divergent target of one of these pairs. At eight further zero-coverage targets the best off-target similarity was 0.967–1.0, and an expanded screen of 1,049 candidates of 60–120~nt across 17 targets added 58 passing probes and no newly covered target. At such loci the appropriate action is to escalate to another assay. A 72-bp TP53 exon, by contrast, was not covered by any passing probe of the 90-nt design. The 70-nt, 35-bp-step design, which also widens the GC and thermodynamic windows, fully covered 10 of 11 TP53 targets, against 9 of 11 in the default design. A GC-rich interval in another target gene (71.9\% GC; 46 of 71 120-bp windows above 70\%) is the candidate for densification. When the same interval was densified in the 69-gene panel described below, 2 of 23 added 120-nt probes passed every thermodynamic threshold, and in a single spike-in experiment with one control and one supplemented library from reference DNA (NA12878)\cite{r78}, unnormalized mean depth over that interval was 2,586× and 3,090×, a single paired observation.

Laboratory capture shows what such a panel achieves. A 69-gene germline panel of 120-nt probes, designed by the probe-design agent, captured 73 human DNA samples in seven batches, with an on-target fraction of 78.3 ± 5.5\% (mean ± s.d.), a median of 99.46\% of target bases at ≥30× and 86 of 87 genotype-derived calls concordant with an orthogonal assay in 29 samples (three calls each; 98.9\%; exact 95\% CI 93.8–99.97\% treating calls as independent) (Fig.~5d). Median on-target fraction ranged from 0.67 to 0.85 across batches, and uniformity also differed between batches. Compared with an initial design written in a single pass by a language-model agent without the structured pipeline, the agent's design of that panel changed exon footprint coverage by more than 10 percentage points in 34 of 69 genes, raising 15 and lowering 19 (Fig.~5e). Its largest gain, C4A/C4B from 4.0\% to 97.2\%, came from shared probes that capture both paralogs and cannot tell them apart, the same limit that the current design meets. The 69-gene design overlapped 111 of the 123 targets of the current design (95 fully, 16 partially).

\begin{figure}[!tbp]
\centering
\includegraphics[width=\textwidth,height=0.72\textheight,keepaspectratio]{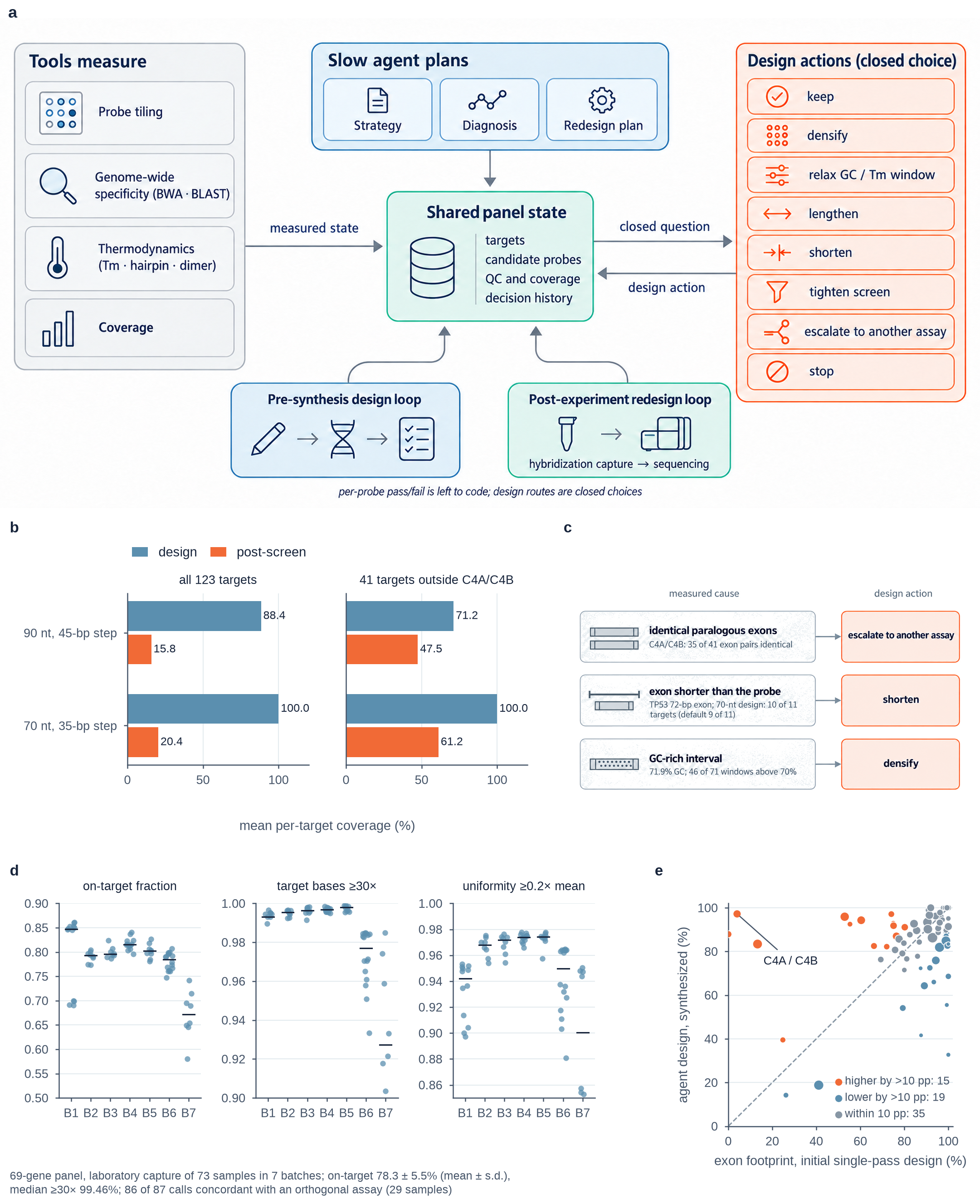}
\caption{\textbf{Germline target enrichment panel design.} \textbf{a}, Division of labour: tools measure, a slow agent plans and handles exceptions in a pre-synthesis design loop and a post-experiment redesign loop, and the remaining decisions are closed choices among eight design actions. \textbf{b}, Mean per-target design and post-screen coverage (equation~(17)) for two designs (90~nt, 45-bp step; 70~nt, 35-bp step), over all 123 targets (82 of which belong to the C4A/C4B pair) and over the 41 targets outside that pair. \textbf{c}, Measured causes of non-coverage and the design actions they map onto: identical paralogous exons (escalate), an exon shorter than a probe (shorten) and a GC-rich interval (densify). \textbf{d}, Laboratory capture of the 69-gene panel: per-sample on-target fraction, fraction of target bases at ≥30× and uniformity in 73 samples from seven batches (B1–B7, chronological); lines, medians; uniformity, fraction of target bases at ≥0.2× the sample's mean depth. \textbf{e}, Exon footprint coverage of the initial single-pass design and of the agent's design that was synthesized and captured, for 69 genes; point size, exon length; orange and blue, change of more than 10 points up or down; grey, within 10 points.}
\label{fig:5}
\end{figure}

\section*{Discussion}

In long-horizon biomedical agents the limit has moved from computation to judgement: most of the original agent's lost items and surplus time came from judgements about conventions, skills, evidence and answer format rather than from the computation. Operon-S1 separates judging from doing and makes these judgements a layer of their own. Each judgement can be checked against a label, the decision model can be exchanged on fixed states, every judgement is traced to its consequence, and the local model answered in under a second at no request cost. With the local model as judge, the layer's efficiency effects were specific to benchmarks, lower cost on SeqQA and fewer steps on protocolqa2, and the rule-based controller also reduced steps on SciAgentArena and seqqa2, which suggests that part of the value lies in posing the right question at the right point. Training on verifiable labels raised the accuracy of every trained version on every decision question, and the gains carried over to the agent's own decision states on held-out items.

The fast decision model here takes the monitoring and control roles that Nelson and Narens\cite{r13} describe. It monitors the state of slower deliberation and controls its course at every recurring judgement, from scoping the task to checking the answer, a role distinct from routing between cheap and expensive models\cite{r17,r32}. Monitoring is measured by how well each judgement matches its label. Control is measured by the interventional contrast of each arm against the no-op control (equation~\eqref{eq:1}). In the original agent's trajectories, headroom differed between judgements (Fig.~2), and the judgements with the most headroom are those that execution can label, which suits a fast, calibrated model that abstains when unsure and learns from execution-derived labels. Protocol troubleshooting has its own map of judgements, and it enters through its own profile of the same harness. Probe design shows the same pattern: per-probe calls reduce to a rule, whereas choices of route, parameters and escalation remain closed decisions over measured state.

The design separates three quantities: whether a decision is right, whether the agent acts on it and whether the task is scored correct. Earlier work found the first and the last to diverge\cite{r36}. The no-op controller anchors the comparison, the provenance record links each decision to its outcome, and the promotion gate admits a new version only on evidence that the training loop does not itself generate (equations~\eqref{eq:3} and \eqref{eq:14}). A scientific agent that distils its trajectories into skills\cite{r79} gains reusable procedures, and the controller adds a measured decision layer.

The operon describes this division of labour: its regulator reads a signal and switches genes without making their products, and a different regulator can be swapped in without changing the genes. The decision layer reads the state of the run, acts only through fixed messages and can be exchanged without changing the agent. The layer is defined by its questions, actions and labels. Six controllers, from rules to a general language model, ran in the same harness on the same states, so the decision model can be exchanged without changing the agent, and a domain enters through a profile of the harness rather than a new architecture. It is likely to help when three conditions hold: (1) a task runs for tens of steps, so that judgements about conventions, routes, evidence and stopping recur; (2) those judgements can be posed as closed questions over a summary of the state; and (3) their correctness can be verified by execution or construction, so that the controller learns without model-generated labels. Computational analysis, laboratory-protocol troubleshooting and probe design all meet these conditions. On short single-step questions the controller's fixed start-up cost dominates, and where a rule reproduces every decision, the rule suffices.

The evaluation covers one slow-system family, its computational work and protocol troubleshooting. The general-LLM arm poses the same closed questions to the model that drives the slow system, which bounds how much of the effect depends on a separate fast model. The protocol profile was tuned on the development half of protocolqa2 and tested once on the held-out half. Skill releases differ between benchmarks, so effects are compared within a benchmark (Methods). In the panel application the current design is computational, and its laboratory reference is the capture of the 69-gene panel.

These properties matter more in the life sciences than in most computational fields. A biological conclusion turns on conventions that a task rarely states, such as the genome build, the coordinate system, the normalization, the threshold and the direction of an effect\cite{r8}, and on attributions such as which step of a protocol failed. A wrong computational judgement costs a rerun, whereas a wrong judgement about a protocol or a probe design costs reagents, instrument time and an experimental cycle. Laboratory practice already answers this with records that let another scientist check each decision, and the provenance record gives an agent's judgements the same form.

Through ProtoPilot's protocol-to-instrument layer\cite{r1}, the same closed decisions, such as whether to densify, shorten or escalate, can be posed during wet-laboratory execution, where their labels come from the experiment itself. As agents move from analysis to instruments, the judgements that decide cost and outcome become a larger share of the work, and a decision layer that judges without doing, and that can be measured, replaced and trained under a gate, is the part of the system that has to be trusted.

\FloatBarrier

\renewcommand{\refname}{References}

\section*{Methods}

\subsubsection*{Slow system}

The slow system is the ProtoPilot agent runtime\cite{r1} (frozen commits 7d92ccb, 7417042 and cb88472, each recorded with its batches) driven by DeepSeek-V4.1-Flash (API model deepseek-flash, build 1017, system fingerprint bc19ba97\ldots{}) with reasoning effort set to high and a 786,432-token context window, beyond which the runtime compacts history. Language-model judges ran at maximum reasoning effort. All arms share one system prompt. Within a benchmark, all arms share one skill release. Release r03 (87 skills, manifest SHA-256 b02db3b6\ldots{}) is used for BiomniBench-DA, SciAgentArena and OmicsBench. Release r04b (92 skills, manifest SHA-256 589b01cd\ldots{}) is used for SeqQA, seqqa2, BixBench-Verified-50 and protocolqa2. r04b contains r03 and adds three things. The first is a general sequence-computation layer: rules for using only the sequences a task provides, six-frame scans of genomic windows and numeric precision, plus skills for melting temperature, molecular weight, sequence composition and complexity, and coding-sequence primer design. The second is a separately named convention layer for LAB-Bench, whose conventions were taken from the published LAB-Bench 2 scoring code and reproduce it on 3,000 synthetic sequences. The third is a patch to the protocol-audit skill. The held-out comparison used r04 and its ablation r04-noB, which removes the convention layer and every reference to it. They differ from r04b and r04b-noB only by the protocol-audit patch. Skills were developed only on the development half of seqqa2 (204 items, by parity of a SHA-256 hash of the item identifier), and the held-out half (196 items) was tested once per release. In a paired ablation on 61 seqqa2 items (41 that an earlier configuration answered correctly and r03 at medium effort did not, and 20 controls), r03 answered 33 correctly at high effort and 31 at medium effort (5 items right only at high effort, 3 only at medium, sign test P = 0.73). With the skills of the earlier configuration, which quoted benchmark wording and were therefore not used, it answered 58 and 54. No effect of the effort setting on seqqa2 was detectable at this sample size (5 against 3 discordant items), whereas the skills of the earlier configuration raised the counts from 33 and 31 to 58 and 54. No release contains item identifiers, sequences, answers or paraphrased items. A leak scan against all 400 seqqa2 items found no sequence, identifier or 25-mer hits, and one 8-gram of generic task wording. Skills are loaded on demand, agents may install packages into a per-task prefix and shared environments are read-only. Per-task time limits are twice the official limit where a benchmark publishes one (SciAgentArena, BiomniBench-DA) and twice the adapter's default otherwise (600~s for SeqQA and seqqa2, 1,200~s for protocolqa2), identical for all arms (Table~2), and timed-out tasks score zero. For seqqa2 the final 60~s of the limit are reserved to collect a last answer, and items answered in this window are labelled as finalized after timeout. A forced final step sends the same explicit final-answer instruction in every arm; it is part of the unmodified agent and therefore of the no-op arm, and its text is identical in every batch of the main table.

\subsubsection*{Motivating trajectories}

The trajectories of Fig.~2 (BixBench-Verified-50, SciAgentArena, seqqa2 and BiomniBench-DA) were recorded with the original agent in earlier rounds, under earlier skill and environment versions, on items that overlap the formal evaluation, and were used only to motivate the design. The values of $\omega$ are single-run, time-weighted descriptive values; for SciAgentArena and seqqa2 the post-answer time is divided by the wall-clock time of all tasks, which bounds $\omega$ from below. The skill-routing counts come from an earlier, smaller skill release. Each wrong BixBench-Verified-50 answer was read in full and assigned to one class. In these trajectories one item, whose reference range has its bounds reversed, could not be graded, so 49 of the 50 items are gradable. In the formal evaluation that range is graded between its two end points in every arm (Benchmarks, deviation (x)).

\subsubsection*{Decision points and states}

A decision point is a tuple
\begin{equation}
d=\Bigl(\mathrm{trig}_d,\ \rho_d,\ \{(q_{d,j},\mathcal{O}_{d,j},\kappa_{d,j})\}_{j\in J_d}\Bigr),
\label{eq:4}
\end{equation}
where $\mathrm{trig}_d(h_t)\in\{0,1\}$ decides whether the point fires on the history $h_t=(a_1,o_1,\ldots,a_t,o_t)$ of agent actions and tool observations, $\rho_d$ renders the history into a state $s=\rho_d(h_t)$, and each sub-question $q_{d,j}$ has a finite option set $\mathcal{O}_{d,j}$ and a type $\kappa_{d,j}\in\{\text{choice},\text{score},\text{yes/no}\}$, with $\mathcal{O}_{d,j}=\{0,1\}$ for yes/no questions and four ordered levels for the score question of the long-computation check. Table~1 lists every decision point. States are rendered by pure functions of the task prompt, workspace listing, environments, skill shortlist and message history; they contain at most 900 tokens (state and questions together at most 1,024), with a fixed task excerpt (the first 1,200 and last 600 characters). Skill candidates are the top 15 BM25\cite{r80} matches over skill descriptions plus a no-skill option. States of a running task contain the last tool call's name, command summary, status, exit code and duration, the tails of standard error (2,000 characters) and standard output (1,000), the budget fraction, loaded skills, a count of repeated commands and the tail of accumulated code (8,000 characters). Runs of 20 or more nucleotide characters are replaced by length placeholders. Option order is shuffled\cite{r81} with a seed derived from the item and decision point and recorded. Conventions come from a registry of slots grouped by analysis type; each slot has a question, options with their meanings, patterns that detect when the task already states the convention and patterns that detect the value the agent used. The registry may not contain an option that is supported only by a proven benchmark defect.

\subsubsection*{Control policy}

A controller $\phi$ returns, for each sub-question, a distribution $p_\phi(\cdot\mid s,q_{d,j})\in\mathcal{P}(\mathcal{O}_{d,j})$, the set of distributions over $\mathcal{O}_{d,j}$; the rule-based and self-verification controllers return point masses and the no-op controller returns nothing. All controllers share the policy, thresholds and messages below and differ only in $p_\phi$, except that the rule-based controller cannot end a run through P7 (see below). A fixed policy maps the distributions and two deterministic state facts, $\mathrm{files}(s)$ and $\mathrm{best}(s)$ (defined below), to an action and a message,
\begin{equation}
\begin{gathered}
\alpha_d=\pi_d\bigl(\{p_\phi(\cdot\mid s,q_{d,j})\}_{j\in J_d},\,s\bigr)\in\{\mathsf{brief},\mathsf{remind},\mathsf{force\text{-}final},\mathsf{recheck},\mathsf{abstain}\},
\\
m=\mu(\alpha_d,\hat y)\in\mathcal{M}\cup\{\varnothing\},
\end{gathered}
\label{eq:5}
\end{equation}
where $\hat y_j=\arg\max_{y}p_\phi(y\mid s,q_{d,j})$, $\mathcal{M}$ is a finite set of templates written in advance and $\varnothing$ denotes no message. Abstention sends no message. Templates may insert the names of files, environments or conventions, never an answer value. The action $\mathsf{force\text{-}final}$ closes the tool loop and sends the final-answer instruction. A choice question yields an action only if its top option is both probable and separated from the runner-up,
\begin{equation}
p_{(1)}\ge\tau_{\text{act}}\quad\text{and}\quad p_{(1)}-p_{(2)}\ge\delta,
\label{eq:6}
\end{equation}
with $\tau_{\text{act}}=0.6$ and $\delta=0.2$. Two decision points can end the tool loop. The evidence check asks whether the evidence answers every part of the task ($p_{\text{done}}$) and whether a required step is missing ($p_{\text{miss}}$); the long-computation check rates four options and uses the probability of stopping with the best result so far ($p_{\text{stop}}$):
\begin{equation}
\begin{gathered}
\pi_{\mathrm{F1}}=
\begin{cases}
\mathsf{force\text{-}final}, & p_{\text{done}}\ge\tau_{\text{done}}\wedge p_{\text{miss}}\le\tau_{\text{miss}}\wedge\neg\,\mathrm{files}(s),\\
\mathsf{remind}, & p_{\text{done}}\ge\tau_{\text{done}}\wedge p_{\text{miss}}\le\tau_{\text{miss}}\wedge\mathrm{files}(s),\\
\mathsf{abstain}, & \text{otherwise},
\end{cases}
\\[4pt]
\pi_{\mathrm{P7}}=
\begin{cases}
\mathsf{force\text{-}final}, & p_{\text{stop}}\ge\tau_{\text{force}}\wedge\mathrm{best}(s)\wedge\phi\neq\mathrm{rule}\wedge\neg\,\mathrm{files}(s),\\
\mathsf{remind}, & p_{\text{stop}}\ge\tau_{\text{stop}}\ \text{and the first case does not apply},\\
\mathsf{abstain}, & \text{otherwise},
\end{cases}
\end{gathered}
\label{eq:7}
\end{equation}
where $\mathrm{files}(s)$ marks tasks answered through output files (BiomniBench-DA and SciAgentArena), for which ending the tool loop would prevent any answer from being written, and $\mathrm{best}(s)$ that the state records a best result so far. The rule-based controller never ends a run through P7, because its stopping signal is a budget fraction rather than a judgement of progress, and it reminds instead. Thresholds were fixed in the decision-layer configuration before any formal benchmark run and are shared by all controllers: $\tau_{\text{format}}=0.3$ (the answer check requests a recheck when the probability that the answer meets its stated format falls below it), $\tau_{\text{act}}=0.6$, $\delta=0.2$, $\tau_{\text{done}}=0.85$, $\tau_{\text{miss}}=0.15$, $\tau_{\text{stop}}=0.8$, $\tau_{\text{force}}=0.9$. The evidence check is asked at most once per task, there is at most one recheck, which requires at least 60~s of remaining budget and is not used for file-answered tasks, and the controller makes at most 60 requests per task, answering every trigger that fires at the same time in a single request. OmicsBench is a multi-turn session benchmark on which the decision layer is inert; it is run in the baseline arm only, and the controller arms run on the other six benchmarks.

\subsubsection*{No-op equivalence and attribution}

Let $\mathrm{req}^{\phi}(h_t)$ be the model request that the agent issues after history $h_t$ under controller $\phi$, and $\mathrm{req}(h_t)$ that of the unmodified agent. On every history in its test suite, a regression test enforces
\begin{equation}
\mathrm{req}^{\phi_0}(h_t)=\mathrm{req}(h_t),
\label{eq:8}
\end{equation}
so that, apart from latency, the agent under the no-op controller behaves as the original agent, and equation~\eqref{eq:1} contrasts interventions that differ only in the controller. Budget fractions in the state are computed from wall-clock time in every arm. The provenance record supports a diagnostic decomposition of $\Delta_\phi$ over decision points,
\begin{equation}
\begin{gathered}
\Delta_\phi=\sum_{d}\Pr[\mathrm{trig}_d]\;\Pr[\alpha_d\neq\mathsf{abstain}\mid\mathrm{trig}_d]\;\Pr[A_d=1\mid\alpha_d\neq\mathsf{abstain},\mathrm{trig}_d]\;\gamma_d\;+\;\xi,
\\
\gamma_d=\mathbb{E}\bigl[Y(\phi)-Y(\phi_0)\mid A_d=1,\mathrm{trig}_d\bigr],
\end{gathered}
\label{eq:9}
\end{equation}
where $A_d$ indicates that the agent adopted the action. The remainder $\xi$ is zero if non-adopted actions have no effect, each decision point fires at most once and the effects of adopted actions at different points add. Because $A_d$ and $\mathrm{trig}_d$ are defined under $\phi$, $\gamma_d$ is not identified from the data, and the adopted-versus-non-adopted contrast we report is descriptive.

\subsubsection*{Post-answer fraction}

For equation~\eqref{eq:2}, time is measured from the start of the task, $t_i^{\ast}$ is the timestamp of the first tool observation whose output contains the submitted value after normalization of number formatting (numbers rounded to the submitted precision, sequences matched as substrings, trivial values not located), and $I$ is the set of tasks for which such an observation exists; for SciAgentArena, $t_i^{\ast}$ is the end of the first run of the solution script that exits successfully. $\omega$ is a ratio of summed times.

\subsubsection*{Monitoring resolution}

The evidence check is posed before the agent's next model request, and the agent's answer almost always arrives in its last message, so in the main run the check fired twice and no evidence-check state had a located answer time. The resolution measure below could therefore not be estimated. Monitoring is evaluated on the states recorded under the no-op controller, which never intervenes, by replaying them through each controller offline. Let $\mathcal{S}^{+}$ and $\mathcal{S}^{-}$ be evidence-check states in tasks that were scored correct, at times $t\ge t_i^{\ast}$ and $t<t_i^{\ast}$ respectively. Both sets come from solved tasks and are separated by whether the answer is already in hand; because the evidence check fires at most once per task, the comparison is between tasks. The monitoring resolution of controller $\phi$, the area under the receiver operating characteristic curve\cite{r82} as used for metacognitive sensitivity\cite{r83}, is the probability that it rates a state in which the answer is already in hand above one in which it is not,
\begin{equation}
\mathrm{AUROC}_\phi=\Pr\bigl[p_{\text{done}}^{\phi}(s^{+})>p_{\text{done}}^{\phi}(s^{-})\bigr]+\tfrac12\Pr\bigl[p_{\text{done}}^{\phi}(s^{+})=p_{\text{done}}^{\phi}(s^{-})\bigr],
\label{eq:10}
\end{equation}
with $s^{+}\in\mathcal{S}^{+}$ and $s^{-}\in\mathcal{S}^{-}$ drawn independently; it is defined when both sets are non-empty. It is 0.5 for a controller that carries no information about the agent's progress. Its calibration on the same states is reported as ECE. Replay uses the deployed per-request timeouts. Intervals are cluster-bootstrap intervals over tasks.

\subsubsection*{Controllers and serving}

All controllers share one client, which sends the state and questions, does not follow redirects, applies per-request time budgets, retries once on rate-limit or server errors and validates the response schema. JEV, LAYA and the general-LLM controller have the same budgets: 30~s for each request, 45~s for each request of the brief, and 92~s for the brief in total. The general-LLM controller sends the same states and questions to DeepSeek-V4.1-Flash at high reasoning effort. A timed-out or failed request leads to abstention and never fails a task. Keys exist only in the orchestrator process and never in the sandbox. The hosted controller calls JEV\cite{r14} (served version jev-1.14.0 on every request). The time budgets of JEV and LAYA were raised to equal those of the general-LLM controller after the main run had started (Deviations during the main run). The local controller calls a server bound to loopback on dedicated CPU cores (four replicas of eight cores behind a local proxy), which agents cannot reach because the egress proxy refuses loopback and private addresses. LAYA version 0 is the public multilingual checkpoint\cite{r47} (revision e4e9ddf); trained versions are registered by name and served by the same process.

\subsubsection*{Synthetic curriculum}

The curriculum has 25 task families covering sequence, alignment, primer, cloning, statistics and omics computations. After removal of four training tasks flagged by a contamination check, it contains 1,517 tasks (1,373 parent tasks and 144 perturbation siblings). Sequences come from public European Nucleotide Archive\cite{r84} references, and sequences that occur inside benchmark items are blacklisted. Each task has two independent implementations and metamorphic checks; before the removal, the verifier accepted every reference answer and rejected all 7,681 constructed negatives, and two independent recomputations, of 100 and of 200 tasks, found no errors (exact two-sided 95\% upper bounds\cite{r85} 3.6\% and 1.8\%; 1.5\% one-sided for the 200 tasks). Splits are grouped by entity, with no entity in more than one split: 972 training, 201 development, 232 test and 112 out-of-distribution tasks. For retraining, about 10\% of training entities, chosen by entity hash, form a calibration split and the rest the training pool; development and test together form the gate split (433 tasks); the out-of-distribution split is read by no selection step. Contamination checks against all benchmarks (8-gram overlap, canary strings, identifiers) found no further hits.

\subsubsection*{Decision synthesis}

Four generators (G1–G3, G5) call the controller's renderer and question functions. G1 labels analysis family and answer type from frozen maps. G2 labels skill relevance from a frozen family-to-skill map, with no-skill rows capped at 25\%. G3 builds compliant and non-compliant final answers in a 1:1 ratio, and compliant answers must pass the task verifier. A fifth generator, G4, builds convention mismatches where a task's constructed negatives map onto a registered slot; it is not used in round 1. G5 injects faults of ten classes into code run in a sandbox (unprivileged user, no network, 20-s timeout, address-space limit) and labels each case with the injected class, kept only when execution fails with that class; in round 1 it ran 10,000 template cases (1,000 per class), discarding cases that exited successfully or duplicated an existing state. Round-1 prose and fault code come from templates. In teacher variants, a language model writes prose around answer and convention placeholders that must be preserved verbatim and writes fault-inducing code; it never sees benchmark text and never produces a label, and its calls are cached by prompt hash and capped in cost. Round 1 produced 12,413 training, 2,561 calibration, 8,281 gate and 1,712 out-of-distribution rows before the firewall (Extended Data Table~1).

\subsubsection*{Leakage firewall}

Every row $r$ with state $s$ carries a label-source tag $\lambda(r)$, a source benchmark, a generator and a lineage that names the batches and arms it derives from. A row enters training only if
\begin{equation}
\begin{aligned}
\mathrm{admit}(r)={}&\mathds{1}[\mathrm{schema}(r)]\;\mathds{1}[\neg\,\mathrm{comp}(r)]\;\mathds{1}[\mathrm{lineage}(r)\ \text{declared}]\;\mathds{1}[\lambda(r)\in\Lambda]\\
&\times\mathds{1}[\mathrm{maxrun}(s)<20]\;\mathds{1}[h(s)\notin H]\;\mathds{1}\bigl[G_8(\mathrm{exc}(s))\cap(G_8(\mathcal{B})\setminus G_8^{\mathrm{bp}})=\varnothing\bigr]=1,
\end{aligned}
\label{eq:11}
\end{equation}
where $\mathrm{schema}$ requires every field to be present, $\mathrm{comp}$ marks lineage through the hosted comparator directly or through any batch that ran a comparator arm, $\Lambda=\{\text{construction},\text{verifier},\text{execution}\}$ (no human-labelled rows exist), $\mathrm{maxrun}$ is the longest run of nucleotide characters, $H$ is the set of hashes of evaluation states, $\mathrm{exc}(s)$ is the task excerpt of the state, $G_8(\mathcal{B})$ is the set of word 8-grams in the task prompts of the six benchmarks evaluated when the round-1 data were built (protocolqa2, added later, is checked against the same rules) and $G_8^{\mathrm{bp}}$ the boilerplate grams of the benchmark prompts that occur in at least $\max(5,\lceil0.02N\rceil)$ of the $N$ prompts. In round 1, with no benchmark states in training, $H$ and $\mathcal{B}$ cover all benchmark items. When benchmark states enter training under cross-fitting, $H$ and $\mathcal{B}$ are restricted to the items of the held-out group. Duplicate state–question pairs are refused at selection. Counts of each rejection class are given in Extended Data Table~1.

\subsubsection*{Hard-example mining}

Within each question, rows are sampled without replacement by taking the largest keys $u_i^{1/w_i}$ with $u_i\sim\mathrm{U}(0,1)$\cite{r86}, with weight
\begin{equation}
w_i=\Bigl(\tfrac12+\tfrac12\,\tfrac{\min(\ell_i,5)}{5}\Bigr)\bigl(1+\mathds{1}[\hat y_i\neq y_i\wedge p_{(1),i}\ge\tau_{\text{act}}]\bigr),
\label{eq:12}
\end{equation}
where $\ell_i$ is the parent version's negative log-likelihood of the label (probabilities floored at $10^{-6}$; $\ell_i=5$ if the parent gave no answer), $\hat y_i$ its top option and $p_{(1),i}$ that option's probability ($\max(p,1-p)$ for yes/no questions). In round 1 the parent is version 0. A total budget of 2,500 rows is allocated across questions by largest remainder. The random-subset control uses the same per-question quotas, sampled uniformly from an independent stream seeded by SHA-256(``13|random'').

\subsubsection*{Training and calibration}

Models were fine-tuned from the public checkpoint with the vendor training tool\cite{r47} on CPU, using soft cross-entropy, shuffled options, maximum length 1,024 tokens (matching the served state budget), seed 13, an effective batch of 64 (micro-batch 1, gradient accumulation 64, 8 threads, identical for candidate and control) and 3 epochs, as an unprivileged user without network access and without swap. Candidate and control were trained under a 12-GiB memory cap with identical selections. Runs with fewer than 100 optimizer updates are refused and training logs are scanned for collapse warnings. A checkpoint's identity is the SHA-256 of its sorted file hashes. Calibration uses temperature scaling\cite{r58}: each question $q$ receives a temperature
\begin{equation}
\beta_q=\arg\min_{\beta\in\mathcal{G}}\ \frac{1}{n_q}\sum_{i\in\mathcal{D}_{\mathrm{cal},q}}-\log\,\mathrm{softmax}\bigl(\log p_i/\beta\bigr)_{y_i},
\qquad \mathcal{G}=\{0.25\cdot16^{j/63}\}_{j=0}^{63},
\label{eq:13}
\end{equation}
with $\beta_q=1$ if $n_q<30$ or if the improvement in negative log-likelihood is at most $10^{-6}$; thresholds are not refitted. The frozen benchmark code serves raw checkpoint probabilities, so each candidate is gated both uncalibrated and calibrated, and only an uncalibrated pass can promote a version into the benchmark arm.

\subsubsection*{Promotion gate}

The gate judges each question kind $q\in\mathcal{Q}$ (relevance questions for different skills are pooled; convention fit is excluded in round 1; rules gate-rules@2, SHA-256 70459d6f\ldots{}), with headroom set $\mathcal{Q}^{\ast}=\{q:\mathrm{acc}_q(\mathrm{rule})\le0.95\}$. With $\widehat{\Delta}_q$ as in equation~\eqref{eq:3}, computed on identical state hashes,
\begin{equation}
\begin{aligned}
\mathrm{promote}(\theta')\iff{}&\mathcal{Q}^{\ast}\neq\varnothing\\
&\;\wedge\;\forall q\in\mathcal{Q}:\ n_q\ge200\ \wedge\ \widehat{\Delta}_q\ge-0.01\ \wedge\ \mathrm{mode}_q(\theta')-\mathrm{mode}_q(y)\le0.20\\
&\;\wedge\;\forall q\in\mathcal{Q}^{\ast}:\ \text{conditions (3)}\\
&\qquad\wedge\ \mathrm{acc}_q(\mathrm{random})\ge\mathrm{acc}_q(\mathrm{majority})-0.05,
\end{aligned}
\label{eq:14}
\end{equation}
where a question that the rule-based controller does not answer counts as incorrect for it, and $\mathrm{mode}_q$ is the share of the most frequent answer and the majority control answers the modal training label. When the modal label share is at least 0.80 the collapse rule cannot bind, and such questions are protected only by the requirement to beat the majority class. Calibration error\cite{r87} is
\begin{equation*}
\mathrm{ECE}_q=\sum_{k=1}^{10}\frac{|B_k|}{n_q}\,\bigl|\mathrm{acc}(B_k)-\mathrm{conf}(B_k)\bigr|
\end{equation*}
over ten equal-width bins of top-option probability; $n_q$ counts the answered gate rows, empty bins contribute zero, and for yes/no questions only bins from 0.5 upwards can be occupied. $Q_{0.025}$ in equation~\eqref{eq:3} is the 2.5th percentile of a percentile cluster bootstrap\cite{r88} (10,000 resamples, seed 13) over entities, or over fault templates for error classes. Verdicts other than promotion are distinguished: insufficient evidence ($n_q<200$), insufficient headroom ($\mathcal{Q}^{\ast}=\varnothing$) and reject. A random-subset control more than 0.05 below the majority class is unhealthy, and the verdict is then insufficient evidence. Control accuracy is computed only on rows that ask the question. A constructed candidate that always answers the most frequent training label, with one-hot or calibrated probabilities, fails the gate in every configuration (20 of 20). An audit found that the first version of the gate could pass a candidate that learned only the label prior, before that version judged any candidate. The revised rules (gate-rules@2, SHA-256 70459d6f\ldots{}) judged every candidate reported here. Training rows of three curriculum families whose prompts contained their own answers were removed before training. For rsi-v6, decision errors of the non-comparator controllers were mined from the graded records of the development halves only (held-out records were skipped before they were opened). The counts per decision type set how many rows each type received, and the rows themselves were new curriculum tasks from fresh generator seeds and re-rendered verified protocols, with labels by construction. The candidate and its random-subset control each drew 1,000 rows from the same pool and shared 200 replay rows (1,200 training rows per arm), the gate split was held out from that pool, and the leakage firewall was applied before either arm was drawn. Offline gate replays use the same 30-s timeout for every controller. Option-order sensitivity was measured by replaying 1,000 gate states with every choice question asked again under a different option order and counting changed answers, with Wilson 95\% intervals. Only the public checkpoint may serve as a root of the registry.

\subsubsection*{Cross-fitting}

For the promoted arm, each benchmark task $x$ is assigned to a group $g(x)=\mathrm{SHA\text{-}256}(\mathrm{key}(x))\bmod K$ with $K=5$; the key is the data capsule for BixBench-Verified-50, the data directory for SciAgentArena and the subtask and item for the others. The promoted arm answers $x$ with
\begin{equation}
\psi(x)=\text{the latest promoted version }\theta\text{ with }g(x)\notin\mathrm{trainedOn}(\theta),
\label{eq:15}
\end{equation}
where $\mathrm{trainedOn}(\theta)$ is the union of the groups on which $\theta$ and all its ancestors were trained; for version 0 it is empty, so $\psi(x)$ always exists. The arm is a mixture of versions. The routing manifest is frozen and hashed before the arm runs. In the main run every task is answered by version 0. Benchmark states entered training only through a parity split of seqqa2 and protocolqa2, the even and odd values of SHA-256 of the item identifier: errors were mined only from the development halves, and in-loop evaluations use only the held-out halves (196 and 62 items).

\subsubsection*{Protocol profile and protocol training data}

The protocol profile is selected per arm and applies per task only when the workspace holds a protocol document (a staged text, PDF or word-processor file whose name or prompt refers to a protocol). Without it, every request and every injected message is byte-identical to the default harness, which a regression test checks. On protocol tasks the evidence, long-computation, error, convention and out-of-bounds checks do not fire. The brief asks at most six questions, among them the domain, the skill, the answer type and the relevance of up to two candidate skills. The answer check requests one recheck when the probability that the answer is specific, or that the change explains the symptom, falls below 0.25. Whether the cited step matches the extracted text is recorded only, because on development transcripts it did not separate correct from incorrect answers. The answer check reads the protocol lines only from the agent's own tool output, and the harness never reads task files itself. The profile arms ran JEV and the general-LLM controller for three rounds each, and LAYA version 0 for two rounds, on protocolqa2 under r04b. For laya-proto, the teacher was Claude Opus 5.5, called through its command-line interface without tools or project context at medium reasoning effort. DeepSeek-V4-Pro, at maximum reasoning effort, wrote the 26 protocols for which Opus returned no structured output. It wrote protocols, faults and facts, and it never produced a label. Each of the eleven domain classes (the ten wet-lab domains and a general class) has seven topic groups with two variants each, 154 protocols in all, and a SHA-256 hash of the group assigns four groups to training, one to calibration and two to the gate. The answer questions of gate and calibration protocols are written to separate held-out files and never enter the gate or calibration rows. Every row carries the lineage of its teacher call and no lineage through the comparator.

\subsubsection*{Comparator}

JEV is used only as an inference-time comparator. Under the provider's terms its outputs are never used as labels, rewards, curricula or selection signals, which the builder, gate and registry enforce through lineage checks. Conclusions about the local controller and its retraining do not depend on it.

\subsubsection*{Benchmarks and deviations from official settings}

SeqQA and seqqa2 report accuracy; BixBench-Verified-50 uses the official open-answer protocol (first answer block, with a language-model judge for range answers); SciAgentArena reports mean correctness, OmicsBench the mean node score and BiomniBench-DA the mean rubric score (0–100). Timeouts score zero. Network policy follows each benchmark: BiomniBench-DA and OmicsBench had web access through a logged proxy that refused benchmark-data hosts, SciAgentArena and protocolqa2 were offline, and SeqQA, seqqa2 and BixBench-Verified-50 used an allowlist of scientific databases that excluded answer sources. Deviations: (i) BiomniBench-DA used the 50 tasks of its public release and ran natively in our sandbox rather than in its container harness, with 2 CPUs and 16~GiB per task; all 50 task images only copy the data directory, which we confirmed. (ii) Its judge script and rubrics were unchanged, but the judge model was DeepSeek-V4-Pro instead of Gemini 3.5 Flash, with maximum output raised from 8,192 to 32,768 tokens because the judge's reasoning otherwise truncated its JSON. (iii) BixBench-Verified-50 range answers were judged by DeepSeek-V4-Pro. (iv) OmicsBench's private test set is unavailable, so its three public tasks were run. (v) SciAgentArena used its 189 publicly reproducible tasks. (vi) BixBench-Verified-50 sets no memory limit; we used 16~GiB and 2 CPUs per task. (vii) Every batch runs on a frozen commit recorded with it. The main table combines two frozen commits that differ only in the time budgets of JEV and LAYA, and rounds started under earlier configurations are excluded. (viii) Time limits are doubled as stated under Slow system. (ix) One R annotation package that would otherwise download gene sets at run time was given a sealed, read-only local cache, identical for all arms. (x) BixBench-Verified-50 item a4d8a810 gives its reference range as (−0.064, −0.084), and we grade it on (−0.084, −0.064). (xi) protocolqa2 publishes no per-task limit, so its limit is twice the adapter's default of 1,200~s, and its open answers are graded with the official labbench2 evaluation prompt by DeepSeek-V4-Pro. Its protocol files are staged from a local copy checked against SHA-256 digests. (xii) When the BiomniBench-DA judge ended a response without any text, the identical request was sent again, at most three times in total and within the verifier time limit. Every request is counted in the judge cost and recorded per task.

\subsubsection*{Deviations during the main run}

The main run started at 01:48 (UTC+8) on 29 October 2026. It departed from its plan in the following ways. Each departure is recorded with the batches it affects, and the analysis rules were fixed before the affected results were computed. All times are UTC+8.

(a) Disk full. Between 01:58 and 02:40 the data volume filled twice. Two BiomniBench-DA tasks staged a full 17.9-GB copy of one count matrix per attempt, and 22 such copies accumulated. The volume was enlarged, and the interrupted batches were resumed under the same identities, configurations and commit. Graded records that showed a disk-full error (313 records, plus 18 that showed it only in the transcript) were not used. Each was replaced by a rerun of the same task under the same configuration, unless its whole batch had already been superseded by a later rerun. No affected task was left without a clean graded record. In a sensitivity analysis, every task with an attempt that started in the 01:58–02:40 window is removed from all arms of its benchmark.

(b) Isolation audit of protocolqa2. The frozen list of answer hosts had no entry for protocolqa2, so its batches were audited with the protocolqa2 hosts added (Evaluation integrity).

(c) Time budgets of JEV and LAYA. At 03:32 the JEV and LAYA batches on BiomniBench-DA and SciAgentArena were stopped at about 20\% completion. They were rerun on a second frozen commit in which JEV and LAYA have the same time budgets as the general-LLM controller, 30~s for each request and 45~s for each request of the brief, instead of the shorter budgets of the first commit. In the main table, JEV and LAYA come from this commit and the other arms from the first. The two commits differ only in these budgets. The stopped batches are kept as evidence for the shorter budgets and are not in the main table.

(d) Concurrency on BiomniBench-DA. After the resume in (a), BiomniBench-DA ran at a concurrency of 3 instead of 6. The baseline, no-op, rule-based, self-verification and general-LLM batches ran partly at 6 and partly at 3, and the JEV and LAYA batches at 3 throughout. Because concurrency changes wall-clock time, the wall-clock and step changes of JEV and LAYA on BiomniBench-DA are shown for reference and do not count towards H1. Their score changes count towards H1.

(e) Host oversubscription and the seqqa2 relaunch. At about 05:30 the host load reached 845 on 96 cores. In the seqqa2 batches running at that point, 185 items had been graded, 36\% to 62\% of them per batch had reached the time limit, and the median wall-clock time per batch was 750 to 1,029~s, against 75~s in the relaunched baseline round. All seqqa2 batches were stopped and relaunched at lower concurrency under new batch identities, with the same configuration, skill release and commit. Only the relaunched batches enter the analysis, and the stopped batches are kept as evidence of the overload. A second SeqQA baseline round stopped at 13 of 600 items when the service manager timed out creating task slices. Its replacement was paused at 137 of 600 items and enters only $\sigma_b$.

(f) Paused baseline rounds. From 05:40, baseline rounds 2–5 on SciAgentArena, BiomniBench-DA and protocolqa2 were paused so that the first round of every arm could finish, and they were resumed later under the same identities. A baseline round enters the reference only if it covers at least 90\% of the items of the most complete round. Extended Data Table~2 reports the number of rounds $R$ that entered on each benchmark. Where $R<5$, the between-round standard deviation and the margin use the actual $R$, and where $R=1$ the margin is undefined and non-inferiority is not judged.

(g) Skill releases. Benchmarks use different releases (Slow system). Effects are reported per benchmark and never pooled across releases. r04b was developed on the seqqa2 development half, so the seqqa2 baseline of the main run may be optimistic. The protocolqa2 main results come from a rerun under r04b that started at 04:06. The first protocolqa2 run, under r03, is reported only as a skill ablation, which also differs in run time and commit.

(h) Protocol profile. The protocol profile of the harness was designed after the first protocolqa2 results were seen and was tuned on a development half of protocolqa2. Its arms, JEV and the general-LLM controller for three rounds each and LAYA version 0 for two rounds, ran on a third frozen commit that adds only the profile. They are reported as exploratory, outside Extended Data Table~2 and outside H1 and H2.

(i) Baseline and no-op. Baseline rounds run the original agent without the harness, and the no-op arm runs in round 1 only. The empirical check of equation~\eqref{eq:8} is therefore the no-op round against the mean of the baseline rounds.

(j) Earlier round. An earlier round at medium reasoning effort is used only for the motivating analyses and the effort comparison. During that round, test suites run on the same host killed tool processes of about 140 live tasks in three windows. These tasks are excluded from every arm of the earlier round. The main run is a new run at high effort, on code in which tests stop only their own units. In the same earlier round, the long-computation check never fired because its timing function rejected the decoded time stamps. Its controller effects therefore come from the other decision points. The main run uses the corrected code.

\subsubsection*{Evaluation integrity}

Each task ran in new network, mount and process namespaces with a private process table, on a minimal root assembled from read-only bind mounts (system directories, the benchmark's environments, reference data and the task's skill snapshot) plus one writable workspace, as an unprivileged user with no capabilities and no privilege escalation. Home, temporary and cache directories were per task, and tasks run offline had no network socket. File tools resolve real paths and refuse writes through symbolic links, and graders that execute agent code run as an unprivileged user in their own process namespace. Before each round, a pre-flight probe had to fail every attempt to read other tasks, earlier rounds, gold answers, grader environments or host files; after each batch, an audit checked for requests that reached an answer host, file-tool paths outside the workspace, the skill snapshot and the agent's own toolchain, new writes to host directories and private sentinels in transcripts and workspaces, and any of these voided the batch. Over the 96 batches of the main table (71 cells and 25 reruns after the disk-full incident; 11,880 records), none met a voiding condition: no request reached an answer host (the proxy blocked 3), and no host write or sentinel hit occurred. Rounds, arms and controller versions carry separate batch identities and cannot be merged or resumed across configurations. Because BiomniBench-DA permits web access, each run was audited after the fact against a table of source studies covering 14 of its 50 tasks; scores are reported over all and over unflagged tasks, and flagged counts are lower bounds. We checked that each language-model judge separates answers of different quality by submitting a strong, a deliberately weakened and an empty answer on three tasks per benchmark. The BiomniBench-DA judge scored them 100/44/0, 100/59/0 and 93/53/0. The BixBench-Verified-50 judge scored 1/0/0 on all three tasks in one run; in another it rejected one strong answer, a verdict it gave once in five gradings of that answer, and we report this variance with its scores. The check exposed a bug in our BiomniBench-DA bridge, which returned an error where the official script scores empty output as zero; it was fixed before these results. Every grader was audited with reproducing scripts. The official seqqa2 validators contain defects, among them water subtracted twice in molecular weight, primer checks on one strand, substring matching of gene names and a unit error in $K_\mathrm{m}$, and the BiomniBench-DA rubric parser drops negative levels. Main tables report official scores unchanged. Every completed batch in the main table, including the resource-failure reruns, passed the isolation audit, with no network request to an answer source and no file access outside the workspace, the skill snapshot and the agent's toolchain.

\subsubsection*{Statistics}

Each arm is paired with the baseline on the same tasks, with the per-task mean of the baseline rounds as reference ($R=5$ scheduled; Extended Data Table~2 gives the rounds that entered). 95\% confidence intervals are cluster-bootstrap intervals over group keys (10,000 resamples, seed 13). Infrastructure failures are excluded and counted. Timeouts keep the official score, which is zero on BiomniBench-DA. Tasks missing from any baseline round or from the arm are excluded and counted. The non-inferiority margin\cite{r89} for benchmark $b$ is
\begin{equation}
\varepsilon_b=\frac{2.8\,\sigma_b}{\sqrt{R}},
\label{eq:16}
\end{equation}
where $\sigma_b$ is the standard deviation of the per-round mean headline score of the baseline over the paired tasks, computed from every baseline round with at least 30 graded tasks, and $R$ is the number of reference rounds (Extended Data Table~2). A benchmark with $R=1$ receives no non-inferiority verdict. The constant 2.8 ($z_{0.975}+z_{0.80}$, for $\alpha=0.05$ and power 0.8) is a power-calculation quantity, and we use it as a margin rather than as a detectable difference. For $R=5$ the margin is $1.25\,\sigma_b$. The margin does not include the run-to-run variation of a single-round arm. A wider margin, $2.8\,\sigma_b\sqrt{1+1/R}$, which includes it, is reported alongside as a sensitivity analysis and is not used for any verdict. H1 holds if the paired relative change in wall-clock time or in steps has an upper bound below zero on at least four of the five hypothesis benchmarks, every hypothesis benchmark with a verdict is non-inferior (lower bound above $-\varepsilon_b$) and controller costs are included. H2 holds if, on SeqQA and seqqa2, JEV beats the self-verification control with a lower bound above zero in score or an upper bound below zero in efficiency. Because H1 and H2 are decision rules over confidence intervals rather than tests with a single statistic, each is judged by its interval conditions. Exact McNemar P values are corrected across benchmarks within each arm by Holm's method\cite{r90}. A hypothesis whose cells are incomplete at the data cutoff is reported as undetermined. Differences are reported in each benchmark's own metric. H3 (which decision point carries the effect) is exploratory, as are all results on protocolqa2 and OmicsBench and all comparisons not named in the hypotheses. A single-round arm whose primary lower bound lies between zero and the margin receives a second round, and the two rounds are judged together, an adaptive design. Intervals resample tasks, and between-round variation of the baseline enters through $\sigma_b$. On SeqQA and seqqa2, later baseline rounds ran at a lower host load than the arms, so wall-clock changes there are paired with baseline round 1, which ran in the same period; scores and steps use all reference rounds. At the data cutoff, arms are paired on the tasks graded in the baseline reference and in every arm of the benchmark, a benchmark enters H1 only if these cover at least max(30, N/2) of its N tasks, and $\sigma_b$ may use the completed part of a paused baseline round with at least 30 tasks. The motivating trajectories used agent commit 1da1a4a and two earlier skill releases (45 and 83 skills).

\subsubsection*{Panel coverage}

For a set $\mathcal{T}$ of targets $g$ and a probe set $P$ with footprints $\mathrm{fp}(p)$, mean per-target design and post-screen coverage and the fully covered fraction are
\begin{equation}
\begin{gathered}
C_{\text{design}}=\frac{1}{|\mathcal{T}|}\sum_{g\in\mathcal{T}}\frac{\bigl|\mathrm{bases}(g)\cap\bigcup_{p\in P}\mathrm{fp}(p)\bigr|}{|\mathrm{bases}(g)|},\\
C_{\text{screen}}=\frac{1}{|\mathcal{T}|}\sum_{g\in\mathcal{T}}\frac{\bigl|\mathrm{bases}(g)\cap\bigcup_{p\in P_{\text{pass}}}\mathrm{fp}(p)\bigr|}{|\mathrm{bases}(g)|},\\
F_{\text{screen}}=\frac{\bigl|\{g\in\mathcal{T}:\mathrm{bases}(g)\subseteq\bigcup_{p\in P_{\text{pass}}}\mathrm{fp}(p)\}\bigr|}{|\mathcal{T}|},
\end{gathered}
\label{eq:17}
\end{equation}
where $P_{\text{pass}}\subseteq P$ are the probes that pass genome-wide specificity screening. For the default design $C_{\text{design}}=0.884$, $C_{\text{screen}}=0.158$ and $F_{\text{screen}}=16/123$; for the 70-nt design $C_{\text{design}}=1.0$, $C_{\text{screen}}=0.204$ and $F_{\text{screen}}=21/123$. Restricted to the 41 targets outside C4A/C4B, $C_{\text{design}}=0.712$ and $C_{\text{screen}}=0.475$ (default) and $1.0$ and $0.612$ (70~nt). The 70-nt design also widens the GC window from 0.30–0.70 to 0.20–0.85 and relaxes the hairpin, dimer and background-hit limits.

\subsubsection*{Probe-design tools}

Probes were tiled with MrBait v1.2.7. Specificity was screened genome-wide on GRCh37 with BWA-MEM v0.7.19 and BLAST 2.16.0 under fixed specificity thresholds, and coverage computed with bedtools v2.31.1\cite{r91} and mosdepth\cite{r92}. Capture libraries were processed with fastp\cite{r93}, BWA-MEM (v0.7.17), Picard MarkDuplicates\cite{r94} and bamdst\cite{r95} against targets padded by 60~bp; uniformity is the fraction of target bases with depth at least 0.2 times the sample's mean depth. Melting temperature, hairpin and self-dimer checks used nearest-neighbour thermodynamics\cite{r69}. Per-probe keep-or-redesign decisions were split by target group into training, development and test sets of 2,818, 827 and 591 records, and the held-out evaluation set is a labelled slice of the test groups with 150 binary decisions (25 targets, 75 of each label). The label is the outcome of the genome-wide specificity screen for the probe. The hosted decision model answered each decision zero-shot as a closed two-option question, and the public LAYA checkpoint was scored on the same set.

\subsubsection*{Compute}

All runs used CPUs only, on a shared 96-core, 182-GB host without GPUs, inside a dedicated control group capped at 90 cores and 160~GB without swap, with separate allocations for benchmark items, controller serving and training. Per-task core budgets were 1 (SeqQA, seqqa2), 2 (BixBench-Verified-50, BiomniBench-DA, protocolqa2) and 4 (SciAgentArena, OmicsBench). Baseline rounds and controller arms ran concurrently on the same host, and absolute wall-clock times depend on its load. Slow-system and judge costs were taken from provider usage and a fixed price table.

\section*{Data availability}

The synthetic curriculum, the decision datasets, the protocol training data, gate reports, decision records and per-task run records are available from Genoria AI. The benchmark items are available from their original releases\cite{r49,r50,r51,r52,r53,r54,r55,r56}.

\section*{Code availability}

The decision harness, controller clients, curriculum and decision-synthesis code, training, calibration and gate code, the evaluation runtime and the analysis code are available from Genoria AI.

\renewcommand{\refname}{Methods references}

\section*{Acknowledgements}

We thank the developers of the benchmarks and open-source tools used in this work.

\section*{Author contributions}

Genoria AI Team designed the study, built the system, ran the experiments, analysed the results and wrote the paper.

\section*{Competing interests}

The authors are employees of Genoria AI.

\FloatBarrier\clearpage

\section*{Extended Data}

\setcounter{table}{0}

\renewcommand{\thetable}{\arabic{table}}\renewcommand{\theHtable}{ED\arabic{table}}

\captionsetup[table]{name=Extended Data Table}

\begin{table}[!htbp]
\centering
\caption{\textbf{Round-1 synthetic decisions.} Question rows per split before the firewall (rows that share a state are counted once per question), gate-split accuracy of the majority class (the most frequent training label; ties broken lexicographically), and gate-split accuracy of the zero-shot checkpoint (version 0) and the rule-based controller after the firewall. Question rows exceed row files because one state can carry several questions. Skill choice is scored against the single constructed label. Under the gate's scoring against each state's set of valid skills, version 0 scores 0.04 and the majority class 0.89 (Fig.~4c).}
\label{edtab:1}
\scriptsize
\setlength{\tabcolsep}{3pt}
\begin{tabularx}{\textwidth}{>{\raggedright\arraybackslash}Xllrrrrrrr}
\toprule
\textbf{Question} & \textbf{Generator} & \textbf{Label source} & \textbf{Train} & \textbf{Cal} & \textbf{Gate} & \textbf{OOD} & \textbf{Majority} & \textbf{Version 0 (gate)} & \textbf{Rule (gate)} \\
\midrule
Analysis family & G1 & construction & 845 & 115 & 428 & 110 & 0.92 & 0.26 & 0.79 \\
Answer type & G1 & construction & 845 & 115 & 428 & 110 & 0.49 & 0.40 & 0.48 \\
Skill choice & G2 & construction & 428 & 115 & 428 & 110 & 0.69 & 0.02 & 0.68 \\
Skill relevance & G2 & construction & 5,892 & 1,550 & 5,739 & 1,430 & 0.87 & 0.13 & 0.05 \\
Answer format & G3 & verifier & 440 & 52 & 212 & 62 & 0.50 & 0.44 & 0.66 \\
Error class & G5 & execution & 4,808 & 729 & 1,474 & 0 & 0.08 & 0.43 & 0.85 \\
\bottomrule
\end{tabularx}
\par\smallskip\begin{minipage}{\textwidth}\footnotesize Row files before the firewall: 12,413 training, 2,561 calibration, 8,281 gate and 1,712 out-of-distribution; the firewall removed 16 calibration and 16 gate rows, leaving 2,545 and 8,265. Accuracies are on the 8,265 gate rows after the firewall (427 rows for analysis family, answer type and skill choice, and 5,725 skill-relevance questions); skill relevance pools all skills.\end{minipage}
\end{table}

\begin{sidewaystable}
\centering
\caption{\textbf{Main results.} Data cutoff 29 October 2026, 18:30 (UTC+8), on the tasks graded in the baseline and in every arm. Official metric of the baseline (mean over the $R$ baseline rounds that entered the reference, ± s.d. across rounds where $R>1$) and, for each controller arm, the paired change against it in score, wall-clock time, agent steps and cost including the controller. Non-inferior: lower bound of the score change above $-\varepsilon_b$ (equation~(16)). H1 is judged on SeqQA, seqqa2, BixBench-Verified-50, SciAgentArena and BiomniBench-DA, and H2 on SeqQA and seqqa2. protocolqa2 is exploratory, and OmicsBench runs the baseline only.}
\label{edtab:2}
\scriptsize
\setlength{\tabcolsep}{4pt}
\renewcommand{\arraystretch}{0.92}
\begin{tabular*}{\linewidth}{@{\extracolsep{\fill}}llrllllll}
\toprule
\textbf{Benchmark} & \textbf{Arm} & \textbf{Tasks} & \textbf{Score} & \textbf{Δ score [95\% CI]} & \textbf{Δ wall (\%)} & \textbf{Δ steps (\%)} & \textbf{Δ cost (\%)} & \textbf{Non-inferior} \\
\midrule
SeqQA & Baseline (R = 2) & 600 & 0.982 ± 0.001 & – & – & – & – & – \\
SeqQA & No-op & 600 & 0.985 & +0.004 [−0.005, +0.013] & −2.3 [−8.2, +3.9] & −0.8 [−5.2, +4.0] & −1.3 [−5.9, +3.5] & no \\
SeqQA & Rule-based & 600 & 0.990 & +0.009 [+0.001, +0.018] & −1.1 [−6.8, +5.1] & −1.4 [−5.8, +3.2] & −3.3 [−8.1, +1.8] & yes \\
SeqQA & Self-verification & 600 & 0.980 & −0.001 [−0.011, +0.008] & −1.1 [−6.9, +5.2] & −2.8 [−7.5, +1.9] & −0.6 [−5.9, +4.9] & no \\
SeqQA & JEV & 600 & 0.982 & +0.001 [−0.009, +0.011] & +1.7 [−4.3, +8.1] & −0.4 [−5.1, +4.6] & +5.9 [+0.0, +12.0] & no \\
SeqQA & LAYA v0 & 600 & 0.985 & +0.004 [−0.005, +0.013] & −4.7 [−10.3, +1.5] & −1.3 [−5.7, +3.3] & −5.8 [−10.8, −0.6] & no \\
SeqQA & General LLM & 600 & 0.987 & +0.006 [−0.003, +0.013] & −2.6 [−8.2, +3.3] & −1.2 [−5.6, +3.4] & +53.6 [+46.4, +61.5] & no \\
seqqa2 & Baseline (R = 3) & 400 & 0.943 ± 0.007 & – & – & – & – & – \\
seqqa2 & No-op & 400 & 0.943 & +0.004 [−0.009, +0.017] & −1.8 [−11.5, +8.1] & −4.5 [−9.1, +0.2] & −3.2 [−10.1, +4.1] & yes \\
seqqa2 & Rule-based & 400 & 0.920 & −0.018 [−0.035, −0.003] & −2.3 [−10.5, +6.1] & −5.5 [−9.8, −0.9] & −7.1 [−12.8, −1.4] & no \\
seqqa2 & Self-verification & 400 & 0.920 & −0.018 [−0.036, −0.002] & −6.0 [−14.2, +2.7] & −2.1 [−6.3, +2.2] & +0.0 [−6.7, +6.8] & no \\
seqqa2 & JEV & 400 & 0.900 & −0.038 [−0.060, −0.018] & +6.9 [−2.2, +16.4] & +2.2 [−2.8, +7.5] & +10.3 [+3.0, +18.1] & no \\
seqqa2 & LAYA v0 & 400 & 0.925 & −0.013 [−0.028, −0.001] & +2.6 [−5.3, +10.8] & +2.1 [−3.0, +7.6] & +6.7 [−1.1, +15.1] & no \\
seqqa2 & General LLM & 400 & 0.890 & −0.048 [−0.072, −0.025] & +8.8 [+0.8, +18.0] & −3.7 [−8.4, +1.3] & +49.8 [+40.5, +60.1] & no \\
BixBench-Verified-50 & Baseline (R = 2) & 50 & 0.880 ± 0.014 & – & – & – & – & – \\
BixBench-Verified-50 & No-op & 50 & 0.840 & −0.030 [−0.092, +0.023] & +2.5 [−10.2, +15.9] & +9.2 [−1.0, +20.0] & +5.0 [−4.3, +16.4] & no \\
BixBench-Verified-50 & Rule-based & 50 & 0.880 & +0.010 [−0.056, +0.064] & −1.0 [−12.8, +9.3] & −5.8 [−15.7, +5.3] & +3.1 [−10.3, +19.5] & no \\
BixBench-Verified-50 & Self-verification & 50 & 0.900 & +0.030 [−0.060, +0.117] & +0.9 [−11.3, +13.7] & +2.2 [−6.8, +10.3] & +7.3 [−2.5, +18.0] & no \\
BixBench-Verified-50 & JEV & 50 & 0.840 & −0.030 [−0.118, +0.044] & +4.2 [−11.2, +21.1] & +4.0 [−6.7, +15.0] & +12.6 [+2.5, +23.8] & no \\
BixBench-Verified-50 & LAYA v0 & 50 & 0.860 & −0.010 [−0.088, +0.056] & −3.3 [−14.4, +9.1] & −3.5 [−11.8, +4.9] & +6.4 [−5.7, +22.3] & no \\
BixBench-Verified-50 & General LLM & 50 & 0.880 & +0.010 [−0.064, +0.077] & −3.0 [−13.3, +7.8] & −4.5 [−13.7, +5.1] & +61.5 [+47.2, +77.0] & no \\
SciAgentArena & Baseline (R = 1) & 189 & 0.655 & – & – & – & – & – \\
SciAgentArena & No-op & 189 & 0.669 & +0.014 [−0.027, +0.049] & −6.9 [−20.2, +5.9] & +0.5 [−10.0, +9.6] & +4.6 [−4.4, +12.1] & – \\
SciAgentArena & Rule-based & 189 & 0.676 & +0.021 [−0.003, +0.043] & −1.8 [−13.6, +12.1] & −9.8 [−19.3, −1.2] & −7.7 [−18.5, +3.5] & – \\
SciAgentArena & Self-verification & 189 & 0.670 & +0.015 [−0.025, +0.048] & −11.0 [−19.0, −0.9] & +3.4 [−6.0, +12.7] & +6.3 [−2.3, +15.0] & – \\
SciAgentArena & JEV & 189 & 0.653 & −0.002 [−0.025, +0.026] & −9.8 [−27.1, +14.2] & −6.2 [−17.7, +4.9] & −2.1 [−15.7, +9.9] & – \\
SciAgentArena & LAYA v0 & 189 & 0.652 & −0.003 [−0.028, +0.029] & −14.0 [−36.9, +6.5] & −6.8 [−18.6, +3.8] & −4.7 [−17.2, +7.6] & – \\
SciAgentArena & General LLM & 189 & 0.659 & +0.004 [−0.025, +0.028] & −5.2 [−14.0, +10.8] & −8.4 [−17.1, −0.2] & +12.9 [+3.9, +22.2] & – \\
BiomniBench-DA & Baseline (R = 2) & 50 & 77.64 ± 0.976 & – & – & – & – & – \\
BiomniBench-DA & No-op & 50 & 79.10 & +0.77 [−2.27, +3.84] & +40.5 [+18.0, +66.4] & −1.3 [−8.5, +6.5] & −1.2 [−8.1, +6.3] & no \\
BiomniBench-DA & Rule-based & 50 & 77.92 & −0.41 [−4.63, +3.88] & −5.5 [−20.2, +12.2] & −6.2 [−14.4, +4.4] & −1.7 [−10.5, +9.7] & no \\
BiomniBench-DA & Self-verification & 50 & 78.10 & −0.23 [−3.86, +3.54] & −10.0 [−24.3, +5.2] & −9.6 [−17.9, +0.6] & −5.6 [−13.8, +5.9] & no \\
BiomniBench-DA & JEV & 50 & 75.78 & −2.55 [−5.73, +0.51] & −18.5 [−31.4, −4.1]† & −16.6 [−23.2, −9.1]† & −15.6 [−22.3, −7.5] & no \\
BiomniBench-DA & LAYA v0 & 50 & 76.06 & −2.27 [−6.25, +1.63] & −14.4 [−26.4, −2.3]† & −13.9 [−21.2, −5.9]† & −7.4 [−14.6, +0.6] & no \\
BiomniBench-DA & General LLM & 50 & 78.18 & −0.15 [−3.72, +3.55] & +16.2 [−2.6, +37.3] & −16.1 [−22.3, −9.5] & −4.4 [−9.9, +1.7] & no \\
protocolqa2 & Baseline (R = 5) & 125 & 0.744 ± 0.021 & – & – & – & – & – \\
protocolqa2 & No-op & 125 & 0.736 & −0.000 [−0.061, +0.061] & −2.2 [−8.9, +4.8] & −3.6 [−8.4, +1.6] & −4.2 [−12.3, +4.1] & no \\
protocolqa2 & Rule-based & 125 & 0.720 & −0.016 [−0.077, +0.045] & −4.6 [−14.1, +5.0] & −6.2 [−13.4, +2.4] & −1.1 [−11.5, +9.7] & no \\
protocolqa2 & Self-verification & 125 & 0.736 & −0.000 [−0.062, +0.062] & −0.6 [−9.2, +8.7] & −1.0 [−7.6, +6.2] & +4.2 [−5.5, +15.0] & no \\
protocolqa2 & JEV & 125 & 0.768 & +0.032 [−0.024, +0.088] & −2.1 [−11.6, +9.4] & −5.1 [−11.6, +2.3] & −1.9 [−11.2, +7.8] & yes \\
protocolqa2 & LAYA v0 & 125 & 0.728 & −0.008 [−0.048, +0.032] & −3.2 [−10.7, +4.4] & −10.3 [−15.7, −4.9] & −0.1 [−9.7, +9.6] & no \\
protocolqa2 & General LLM & 125 & 0.736 & −0.000 [−0.058, +0.058] & −2.0 [−10.6, +7.1] & −10.9 [−16.8, −4.8] & +8.6 [−0.9, +19.1] & no \\
OmicsBench & Baseline (R = 5) & 3 & 0.861 ± 0.051 & – & – & – & – & – \\
\bottomrule
\end{tabular*}
\par\smallskip\begin{minipage}{\textwidth}\footnotesize Score: accuracy (SeqQA, seqqa2, protocolqa2), open-answer accuracy (BixBench-Verified-50), mean correctness (SciAgentArena), rubric score 0–100 (BiomniBench-DA) and mean node score (OmicsBench). Changes in wall-clock time, steps and cost are ratios of paired sums minus one. On SeqQA and seqqa2, wall-clock changes are paired with baseline round 1, which ran in the same period as the arms (Methods). Cost counts agent tokens and controller requests and excludes grading. On BiomniBench-DA, the wall-clock and step changes of JEV and LAYA (†) are for reference only, because their batches ran at a different concurrency (Methods). $\varepsilon_b$ uses the baseline rounds with at least 30 graded tasks (Methods); SciAgentArena has one reference round (R = 1), and its non-inferiority is not judged (–).\end{minipage}
\end{sidewaystable}

\begin{table}[!htbp]
\centering
\caption{\textbf{No-op control against the baseline.} Paired change of the no-op arm against the baseline in official score and wall-clock time, the between-round standard deviation of the baseline, the share of tasks whose outcome changed between baseline rounds, and the audit of injected messages for gold answers, per benchmark. No injected message contained the gold answer in the 2,361 runs whose answers can be matched as text.}
\label{edtab:3}
\footnotesize
\setlength{\tabcolsep}{4pt}
\begin{tabularx}{\textwidth}{lr>{\raggedright\arraybackslash}X>{\raggedright\arraybackslash}X>{\raggedright\arraybackslash}X>{\raggedright\arraybackslash}X>{\raggedright\arraybackslash}X}
\toprule
\textbf{Benchmark} & \textbf{Tasks} & \textbf{Δ score [95\% CI]} & \textbf{Δ wall (\%)} & \textbf{Δ steps (\%)} & \textbf{Baseline σ (rounds used for σ)} & \textbf{Tasks changed between baseline rounds} \\
\midrule
SeqQA & 600 & +0.004 [−0.005, +0.013] & −2.3 [−8.2, +3.9] & −0.8 [−5.2, +4.0] & 0.001 (2) & 2.2\% \\
seqqa2 & 400 & +0.004 [−0.009, +0.017] & −1.8 [−11.5, +8.1] & −4.5 [−9.1, +0.2] & 0.007 (3) & 3.0\% \\
protocolqa2 & 125 & −0.000 [−0.061, +0.061] & −2.2 [−8.9, +4.8] & −3.6 [−8.4, +1.6] & 0.021 (5) & 13.1\% \\
BixBench-Verified-50 & 50 & −0.030 [−0.092, +0.023] & +2.5 [−10.2, +15.9] & +9.2 [−1.0, +20.0] & 0.014 (2) & 10.0\% \\
SciAgentArena & 189 & +0.014 [−0.027, +0.049] & −6.9 [−20.2, +5.9] & +0.5 [−10.0, +9.6] & 0.010 (5) & 8.5\% \\
BiomniBench-DA & 50 & +0.770 [−2.270, +3.840] & +40.5 [+18.0, +66.4] & −1.3 [−8.5, +6.5] & 0.979 (3) & 0.0\% \\
\bottomrule
\end{tabularx}
\end{table}

\begin{table}[!htbp]
\centering
\caption{\textbf{Tasks gained and lost against the baseline.} For each controller arm and benchmark, tasks right in both, gained (wrong to right), lost (right to wrong) and wrong in both, with exact McNemar P values, Holm-corrected across benchmarks. BiomniBench-DA has no binary correctness and is not listed.}
\label{edtab:4}
\scriptsize
\setlength{\tabcolsep}{3pt}
\begin{tabularx}{\textwidth}{>{\raggedright\arraybackslash}Xlrrrrrrr}
\toprule
\textbf{Arm} & \textbf{Benchmark} & \textbf{Tasks} & \textbf{Right in both} & \textbf{Gained} & \textbf{Lost} & \textbf{Wrong in both} & \textbf{P} & \textbf{P (Holm)} \\
\midrule
No-op & SeqQA & 600 & 584 & 7 & 5 & 4 & 0.774 & 1.000 \\
No-op & seqqa2 & 400 & 375 & 2 & 5 & 18 & 0.453 & 1.000 \\
No-op & protocolqa2 & 125 & 82 & 10 & 12 & 21 & 0.832 & 1.000 \\
No-op & BixBench-Verified-50 & 50 & 42 & 0 & 2 & 6 & 0.500 & 1.000 \\
No-op & SciAgentArena & 189 & 101 & 8 & 7 & 73 & 1.000 & 1.000 \\
Rule-based & SeqQA & 600 & 586 & 8 & 3 & 3 & 0.227 & 0.906 \\
Rule-based & seqqa2 & 400 & 368 & 0 & 12 & 20 & 0.000 & 0.002 \\
Rule-based & protocolqa2 & 125 & 82 & 8 & 12 & 23 & 0.503 & 1.000 \\
Rule-based & BixBench-Verified-50 & 50 & 43 & 1 & 1 & 5 & 1.000 & 1.000 \\
Rule-based & SciAgentArena & 189 & 99 & 12 & 9 & 69 & 0.664 & 1.000 \\
Self-verification & SeqQA & 600 & 582 & 6 & 7 & 5 & 1.000 & 1.000 \\
Self-verification & seqqa2 & 400 & 366 & 2 & 14 & 18 & 0.004 & 0.021 \\
Self-verification & protocolqa2 & 125 & 83 & 9 & 11 & 22 & 0.824 & 1.000 \\
Self-verification & BixBench-Verified-50 & 50 & 42 & 3 & 2 & 3 & 1.000 & 1.000 \\
Self-verification & SciAgentArena & 189 & 103 & 8 & 5 & 73 & 0.581 & 1.000 \\
JEV & SeqQA & 600 & 583 & 6 & 6 & 5 & 1.000 & 1.000 \\
JEV & seqqa2 & 400 & 359 & 1 & 21 & 19 & 0.000 & 0.000 \\
JEV & protocolqa2 & 125 & 85 & 11 & 9 & 20 & 0.824 & 1.000 \\
JEV & BixBench-Verified-50 & 50 & 41 & 1 & 3 & 5 & 0.625 & 1.000 \\
JEV & SciAgentArena & 189 & 100 & 7 & 8 & 74 & 1.000 & 1.000 \\
LAYA v0 & SeqQA & 600 & 586 & 5 & 3 & 6 & 0.727 & 1.000 \\
LAYA v0 & seqqa2 & 400 & 370 & 0 & 10 & 20 & 0.002 & 0.010 \\
LAYA v0 & protocolqa2 & 125 & 89 & 2 & 5 & 29 & 0.453 & 1.000 \\
LAYA v0 & BixBench-Verified-50 & 50 & 42 & 1 & 2 & 5 & 1.000 & 1.000 \\
LAYA v0 & SciAgentArena & 189 & 102 & 7 & 6 & 74 & 1.000 & 1.000 \\
General LLM & SeqQA & 600 & 586 & 6 & 3 & 5 & 0.508 & 1.000 \\
General LLM & seqqa2 & 400 & 355 & 1 & 25 & 19 & 0.000 & 0.000 \\
General LLM & protocolqa2 & 125 & 84 & 8 & 10 & 23 & 0.815 & 1.000 \\
General LLM & BixBench-Verified-50 & 50 & 42 & 2 & 2 & 4 & 1.000 & 1.000 \\
General LLM & SciAgentArena & 189 & 98 & 7 & 10 & 74 & 0.629 & 1.000 \\
\bottomrule
\end{tabularx}
\end{table}

\begin{table}[!htbp]
\centering
\caption{\textbf{Protocol profile on protocolqa2 (exploratory).} JEV, the general-LLM controller and LAYA version 0 with the protocol profile against the baseline and against the same controllers under the default harness, in score, wall-clock time and steps. Protocol-profile arms are averaged per item over their rounds: three for JEV and the general-LLM controller, two for LAYA version 0, whose wall-clock change is omitted because its batches ran in a different load period.}
\label{edtab:5}
\footnotesize
\setlength{\tabcolsep}{4pt}
\begin{tabularx}{\textwidth}{>{\raggedright\arraybackslash}Xrlll}
\toprule
\textbf{Comparison} & \textbf{Tasks} & \textbf{Δ score [95\% CI]} & \textbf{Δ wall (\%)} & \textbf{Δ steps (\%)} \\
\midrule
JEV + protocol profile − baseline & 125 & +0.011 [−0.023, +0.045] & −11.7 [−17.8, −5.0] & −4.2 [−9.2, +1.0] \\
General LLM + protocol profile − baseline & 125 & +0.035 [−0.002, +0.073] & −15.0 [−21.3, −8.4] & −9.7 [−14.1, −5.4] \\
JEV + protocol profile − JEV (default harness) & 125 & −0.021 [−0.077, +0.035] & −9.8 [−18.7, −0.2] & +0.9 [−5.8, +7.8] \\
General LLM + protocol profile − general LLM (default harness) & 125 & +0.035 [−0.024, +0.093] & −13.3 [−21.3, −4.6] & +1.4 [−5.5, +8.5] \\
LAYA version 0 + protocol profile − baseline & 125 & +0.060 [+0.014, +0.108] & – & −7.9 [−12.3, −3.3] \\
LAYA version 0 + protocol profile − baseline, held-out half & 62 & +0.055 [−0.011, +0.129] & – & −8.9 [−15.5, −1.3] \\
LAYA version 0 + protocol profile − LAYA version 0 (default harness) & 125 & +0.068 [+0.012, +0.128] & – & +2.6 [−3.7, +9.4] \\
\bottomrule
\end{tabularx}
\end{table}

\begin{table}[!htbp]
\centering
\caption{\textbf{Decision accuracy on benchmark states.} Main-table round 1 of the rule-based, LAYA version 0, JEV and general-LLM arms on SeqQA, seqqa2, protocolqa2, BixBench-Verified-50, SciAgentArena and BiomniBench-DA, original batches graded by 29 October 2026, 11:00 (UTC+8). Labels are execution-derived (error class, from the failing call's output), benchmark-derived (analysis family, answer type) or a grader proxy (answer format: the grader can parse the answer). Rule accuracy counts an unanswered state as wrong, as the gate does. Model columns give accuracy [95\% cluster-bootstrap interval] / expected calibration error (answered states). n, labelled states (answer format: states at which it was asked). Headroom: rule accuracy at most 0.95; for error class it holds on BixBench-Verified-50, SciAgentArena and BiomniBench-DA. The answer-type label exists only on SeqQA and has one value. Questions without a model-free label on benchmark states are not listed. General-LLM values use original batches only and differ by at most 0.004 from Extended Data Fig.~3, which includes resource-failure reruns.}
\label{edtab:6}
\scriptsize
\setlength{\tabcolsep}{3pt}
\begin{tabularx}{\textwidth}{lrrr>{\raggedright\arraybackslash}X>{\raggedright\arraybackslash}X>{\raggedright\arraybackslash}Xl}
\toprule
\textbf{Question} & \textbf{n} & \textbf{Majority} & \textbf{Rule} & \textbf{LAYA v0} & \textbf{JEV} & \textbf{General LLM} & \textbf{Headroom} \\
\midrule
Error class (P3) & 3,434 & 0.383 & 0.953 & 0.164 [0.132, 0.194] / 0.360 (836) & 0.864 [0.841, 0.885] / 0.036 (868) & 0.846 [0.820, 0.871] / 0.026 (891) & on 3 of 6 benchmarks \\
Analysis family (S1) & 4,554 & 0.878 & 0.593 & 0.087 [0.052, 0.126] / 0.399 (1,139) & 0.974 [0.961, 0.984] / 0.013 (1,139) & 0.963 [0.944, 0.980] / 0.040 (941) & yes \\
Answer type (S7) & 2,400 & 1.000 & 0.933 & 0.663 [0.625, 0.702] / 0.267 (600) & 1.000 [1.000, 1.000] / 0.039 (600) & 0.998 [0.994, 1.000] / 0.046 (472) & yes (one label value) \\
Answer format (D20) & 481 & 0.996 & 0.996 & 0.994 [0.985, 1.000] / 0.053 (477) & 0.680 [0.638, 0.722] / 0.132 (485) & 0.903 [0.875, 0.928] / 0.054 (473) & no \\
\bottomrule
\end{tabularx}
\end{table}

\FloatBarrier

\noindent\includegraphics[width=\textwidth,height=0.6\textheight,keepaspectratio]{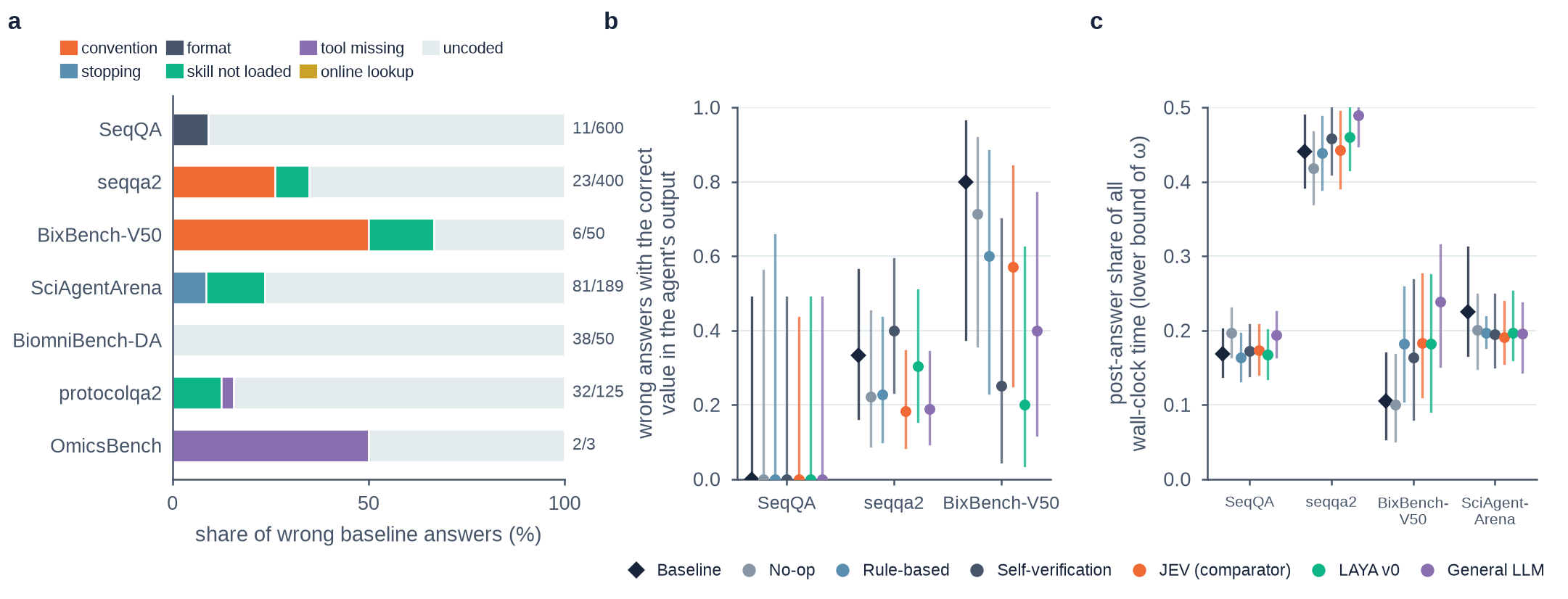}\par\smallskip

\noindent{\small\textbf{Extended Data Fig. 1 $|$ Errors after the computation in the formal evaluation.} \textbf{a}, Failure classes of wrong baseline answers by rule codebook, per benchmark (wrong/total: SeqQA 11/600, seqqa2 23/400, BixBench-Verified-50 6/50, SciAgentArena 81/189, BiomniBench-DA 38/50 below full marks, protocolqa2 32/125, OmicsBench 2/3). Convention errors are detectable only on benchmarks with answer matchers (SeqQA, seqqa2 and BixBench-Verified-50). \textbf{b}, Share of wrong answers with an answer matcher whose correct value had appeared in the agent's own tool output, per arm, with Wilson 95\% intervals. \textbf{c}, Post-answer share of all wall-clock time, a lower bound on $\omega$, per arm and benchmark, with 95\% cluster-bootstrap intervals.\par}\bigskip

\noindent\includegraphics[width=\textwidth,height=0.6\textheight,keepaspectratio]{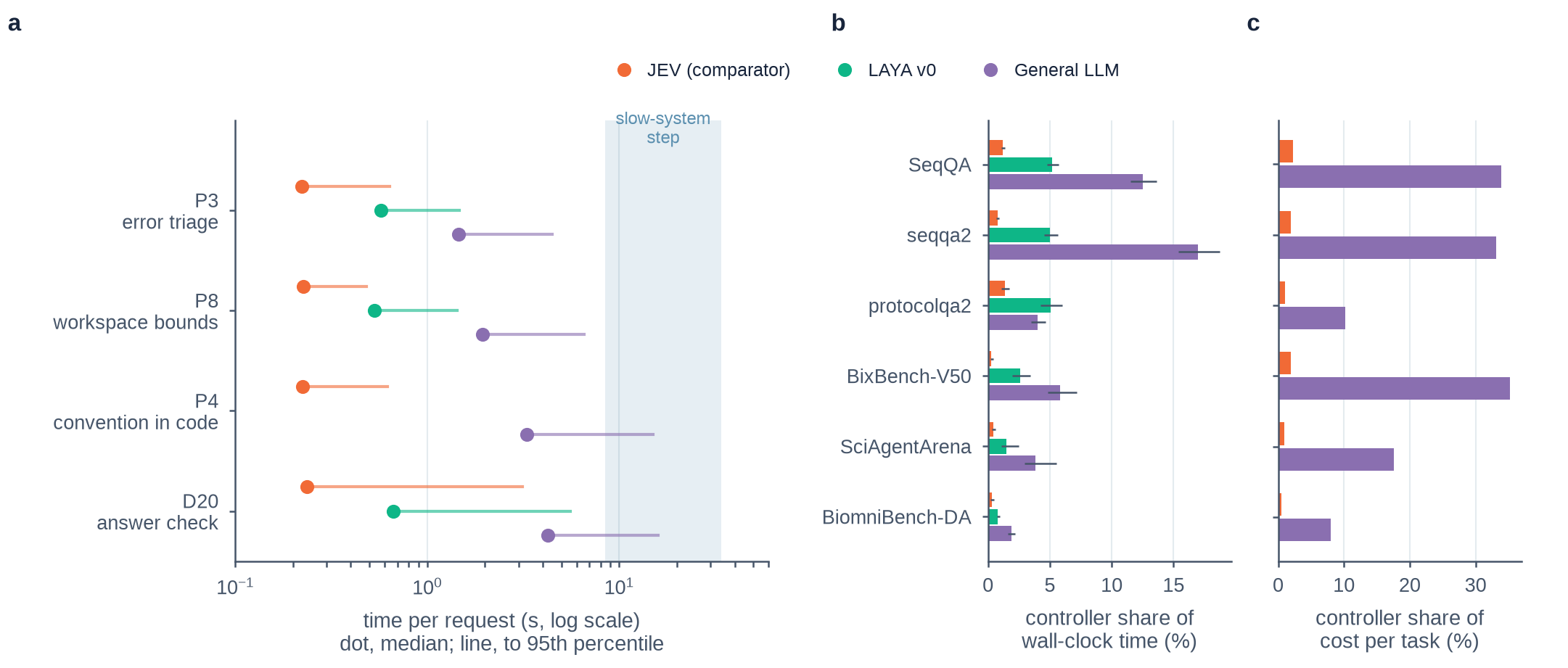}\par\smallskip

\noindent{\small\textbf{Extended Data Fig. 2 $|$ Speed and cost including the controller.} \textbf{a}, Time per controller request at four decision points of the formal evaluation (Table~1), for LAYA version 0 on CPU, JEV and the general-LLM controller. Dots are medians and lines extend to the 95th percentile. Points with fewer than 30 requests are omitted. The shaded band is the range of the median slow-system step across benchmarks, computed as wall-clock time minus controller time, divided by agent steps, over the baseline rounds of the main table. \textbf{b}, Controller share of wall-clock time per benchmark for JEV, LAYA version 0 and the general-LLM controller, with 95\% intervals. \textbf{c}, Controller share of cost per task for JEV and the general-LLM controller. The brief is not shown. LAYA runs locally with no per-request price.\par}\bigskip

\noindent\includegraphics[width=\textwidth,height=0.6\textheight,keepaspectratio]{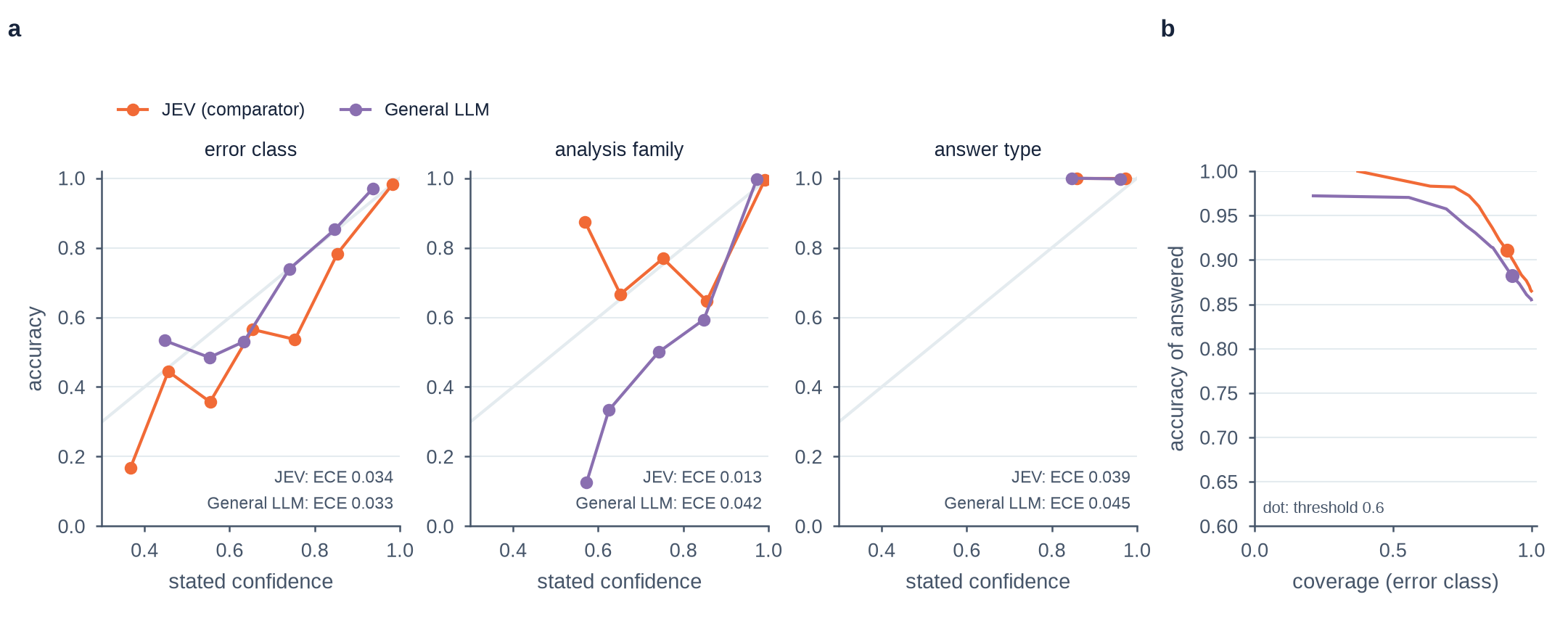}\par\smallskip

\noindent{\small\textbf{Extended Data Fig. 3 $|$ Calibration and selective prediction.} \textbf{a}, Reliability diagrams of JEV and the general-LLM controller on benchmark states, for the error class (labels from the failing call's output), the analysis family and the answer type (labels from the benchmark). Bins of 0.1 in stated confidence with at least five decisions are shown, and each panel gives the expected calibration error. \textbf{b}, Accuracy of answered error-triage questions against coverage as the action threshold varies. Dots mark the threshold of 0.6 used in the evaluation.\par}\bigskip

\noindent\includegraphics[width=\textwidth,height=0.6\textheight,keepaspectratio]{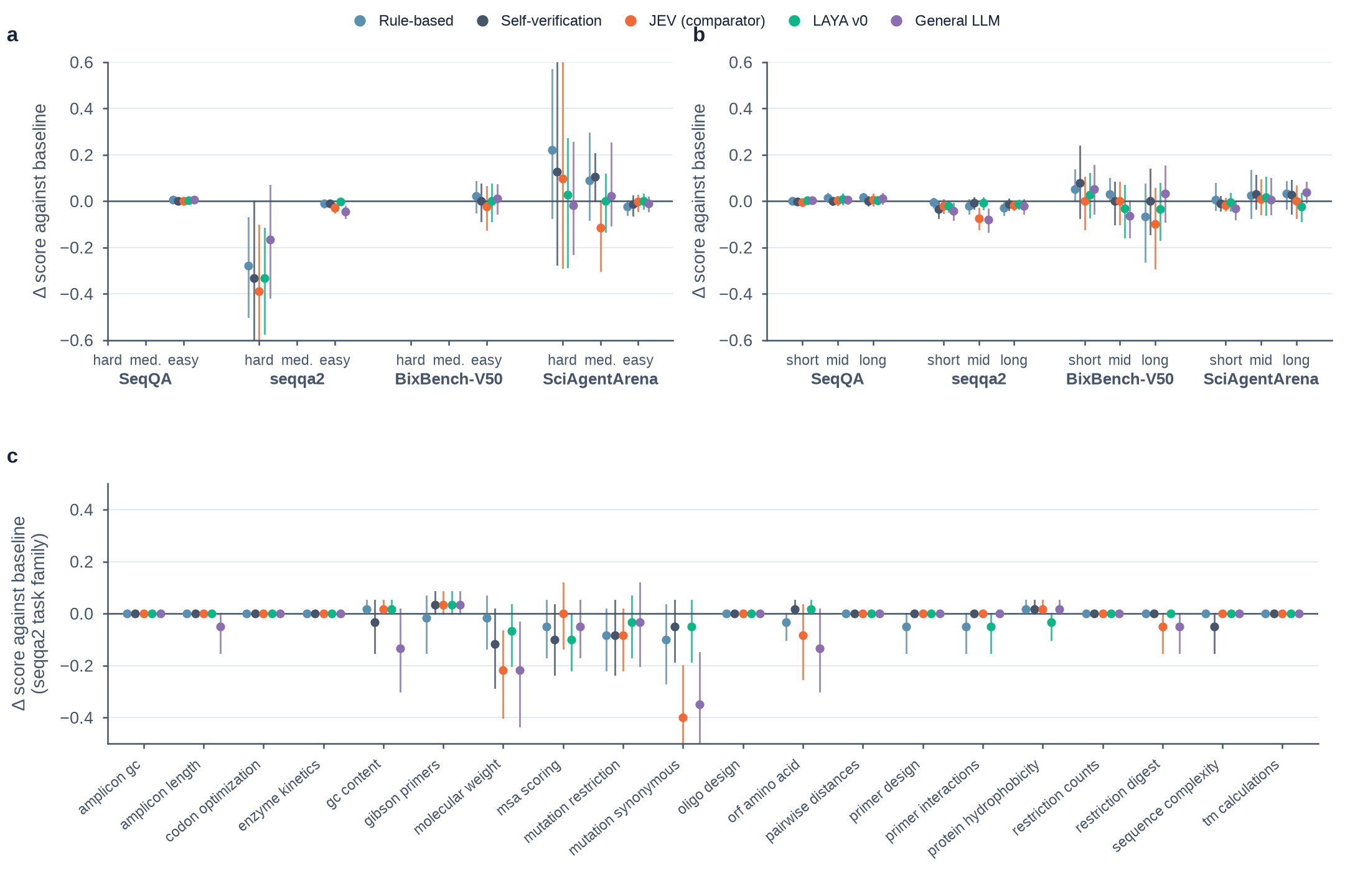}\par\smallskip

\noindent{\small\textbf{Extended Data Fig. 4 $|$ Where the controller helps.} Paired score change against the baseline by difficulty stratum (pass rate over the other controller arms) (\textbf{a}), by tertile of trajectory length in agent steps (\textbf{b}) and per seqqa2 task family, 15–20 tasks per family (\textbf{c}). Strata with fewer than ten tasks are not shown. Error bars, 95\% cluster-bootstrap intervals, clipped at the axis limits.\par}\bigskip

\noindent\includegraphics[width=\textwidth,height=0.6\textheight,keepaspectratio]{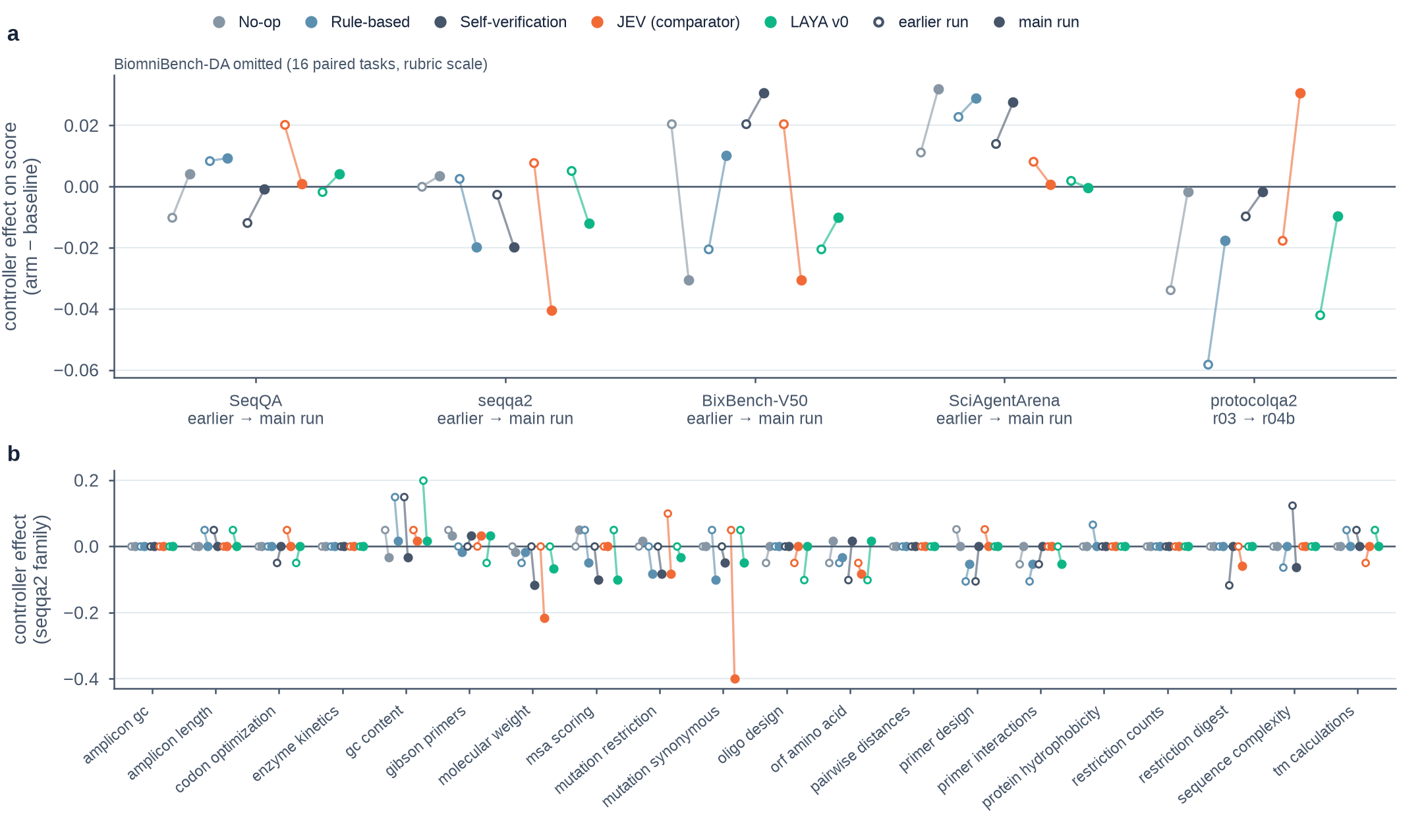}\par\smallskip

\noindent{\small\textbf{Extended Data Fig. 5 $|$ Skill release and controller effect.} \textbf{a}, Controller effect on score (arm minus baseline) in the earlier run (hollow) and in the main run (filled), per benchmark and arm (paired tasks: SeqQA 594, seqqa2 386, BixBench-Verified-50 49, SciAgentArena 158). The two runs differ in skill release and in the agent's reasoning effort. For protocolqa2 (124 tasks) the comparison is skill release r03 against r04b at the same effort, and the two series also differ in frozen code and run time. The general-LLM arm did not run in the earlier run, and BiomniBench-DA is omitted (16 paired tasks). \textbf{b}, The same comparison per seqqa2 validator family (15–20 tasks per family).\par}\bigskip

\end{document}